\documentstyle[11pt,fullname,a4wide]{article}
\input{epsf}

\begin{document}

\title{Forgetting Exceptions is Harmful in Language
Learning\thanks{This is a preprint version of an article that will appear
in {\em Machine Learning}, {\bf 11:1--3}, pp. 11--42.}}

\author{Walter Daelemans, Antal van den Bosch, Jakub Zavrel \\
ILK / Computational Linguistics \\
Tilburg University \\
P.O. Box 90153 \\
NL-5000 LE Tilburg \\
The Netherlands \\
e-mail \{walter,antalb,zavrel\}@kub.nl}

\date{}
\maketitle

\abstract{We show that in language learning, contrary to received
wisdom, {\em keeping exceptional training instances in memory}\/ can be
beneficial for generalization accuracy. We investigate this phenomenon
empirically on a selection of benchmark natural language processing
tasks: grapheme-to-phoneme conversion, part-of-speech tagging,
prepositional-phrase attachment, and base noun phrase chunking. In a
first series of experiments we combine memory-based learning with
training set editing techniques, in which instances are edited based
on their typicality and class prediction strength. Results show that
editing exceptional instances (with low typicality or low class
prediction strength) tends to harm generalization accuracy. In a
second series of experiments we compare memory-based learning and
decision-tree learning methods on the same selection of tasks, and
find that decision-tree learning often performs worse than
memory-based learning. Moreover, the decrease in performance can be
linked to the degree of abstraction from exceptions (i.e., pruning or
eagerness). We provide explanations for both results in terms of the
properties of the natural language processing tasks and the learning
algorithms.}

\section{Introduction}

Memory-based reasoning \cite{Stanfill+86} is founded on the hypothesis
that performance in real-world tasks (in our case language processing)
is based on reasoning on the basis of similarity of new situations to
{\em stored representations of earlier experiences}, rather than on
the application of {\em mental rules}\/ abstracted from earlier
experiences as in rule-based processing.  The type of learning
associated with such an approach is called {\em lazy learning}\/
\cite{Aha97a}.  The approach has surfaced in different contexts using
a variety of alternative names such as example-based, exemplar-based,
analogical, case-based, in\-stance-based, locally weighted, and
memory-based \cite{Stanfill+86,Cost+93,Kolodner93,Aha+91,Atkeson+97}.
Historically, lazy learning algorithms are descendants of the
$k$-nearest neighbor (henceforth $k$-{\sc nn}) classifier
\cite{Cover+67,Devijver+82,Aha+91}.

Memory-based learning is `lazy' as it involves adding training
examples (feature-value vectors with associated categories) to memory
without abstraction or restructuring. During classification, a
previously unseen test example is presented to the system. Its
similarity to all examples in memory is computed using a {\em
similarity metric}, and the category of the most similar example(s) is
used as a basis for extrapolating the category of the test example.  A
key feature of memory-based learning is that, normally, {\em all}\/
examples are stored in memory and no attempt is made to simplify the
model by eliminating noise, low frequency events, or exceptions.
Although it is clear that noise in the training data can harm accurate
generalization, this work focuses on the problem that, for language
learning tasks, it is very difficult to discriminate between noise on
the one hand, and valid exceptions and sub-regularities that are
important for reaching good accuracy on the other hand.

The goal of this paper is to provide empirical evidence that for a
range of language learning tasks, memory-based learning methods tend
to achieve better generalization accuracies than (i) memory-based
methods combined with training set editing techniques in which
exceptions are explicitly forgotten, i.e.~removed from memory, and
(ii) decision-tree learning in which some of the information from the
training data is either forgotten (by pruning) or made inaccessible
(by the eager construction of a model).  We explain these results in
terms of the data characteristics of the tasks, and the properties of
memory-based learning. In our experiments we compare {\sc ib1-ig}
\cite{Daelemans+92b,Daelemans+97}, a memory-based learning algorithm,
with (i) edited versions of {\sc ib1-ig}, and (ii) decision-tree
learning in {\sc c5.0} \cite{Quinlan93} and in {\sc igtree}
\cite{Daelemans+97}. These learning methods are described in
Section~\ref{methods}. The compared algorithms are applied to a
selection of four natural language processing ({\sc nlp}) tasks
(described in Section \ref{tasks}). These tasks present a varied
sample of the complete domain of {\sc nlp} as they relate to phonology
and morphology (grapheme-to-phoneme conversion); morphology and syntax
(part of speech tagging, base noun phrase chunking); and syntax and
lexical semantics (prepositional-phrase attachment).

First, we show in Section~\ref{editing} that two criteria for editing
instances in memory-based learning, viz. low typicality and low class
prediction strength, are generally responsible for a decrease in
generalization accuracy. 

Second, memory-based learning is demonstrated in Section~\ref{dectree}
to be mostly at an advantage, and sometimes at a par with
decision-tree learning as far as generalization accuracy is concerned.
The advantage is puzzling at first sight, as {\sc ib1-ig}, {\sc c5.0}
and {\sc igtree} are based on similar principles: (i) classification
of test instances on the basis of their similarity to training
instances (in the form of the instances themselves in {\sc ib1-ig} or
in the form of hyper-rectangles containing subsets of partly-similar
training instances in {\sc c5.0} and {\sc igtree}), and (ii) use of
information entropy as a heuristic to constrain the space of possible
generalizations (as a feature weighting method in {\sc ib1-ig}, and as
a split criterion in {\sc c5.0} and {\sc igtree}).

Our hypothesis is that both effects are due to the fact that {\sc
ib1-ig} keeps all training instances as possible sources for
classification, whereas both the edited versions of {\sc ib1-ig} and
the decision-tree learning algorithms {\sc c5.0} and {\sc igtree} make
abstractions from irregular and low-frequency events.  In language
learning tasks, where sub-regularities and (small families of)
exceptions typically abound, the latter is detrimental to generalization
performance.  Our results suggest that forgetting exceptional training
instances is harmful to generalization accuracy for a wide range of
language-learning tasks.  This finding contrasts with a consensus in
supervised machine learning that forgetting exceptions by pruning
boosts generalization accuracy \cite{Quinlan93}, and with studies
emphasizing the role of forgetting in learning
\cite{Markovitch+88,Salganicoff93}.

Section \ref{why} places our results in a broader machine learning and
language learning context, and attempts to describe the properties of
language data and memory-based learning that are responsible for the
`forgetting exceptions is harmful' effect.  For our data sets, the
abstraction and pruning techniques studied do not succeed in reliably
distinguishing noise from productive exceptions, an effect we
attribute to a special property of language learning tasks: the
presence of many exceptions that tend to occur in groups or {\em
pockets}\/ in instance space, together with noise introduced by corpus
coding methods.  In such a situation, the best strategy is to keep all
training data to generalize from.

\section{Learning methods}
\label{methods}

In this Section, we describe the three algorithms we used in our 
experiments.  {\sc ib1-ig} is used for studying the effect of editing 
exceptional training instances, and in a comparison to the decision 
tree methods {\sc c5.0} and {\sc igtree}.

\subsection{IB1-IG}
\label{ib1ig}

{\sc ib1-ig} \cite{Daelemans+92b,Daelemans+97} is a memory-based (lazy)
learning algorithm that builds a data base of instances (the {\em
instance base}) during learning. An instance consists of a
fixed-length vector of $n$ feature-value pairs, and a field containing
the classification of that particular feature-value vector. After the
instance base is built, new (test) instances are classified by
matching them to all instances in the instance base, and by
calculating with each match the {\em distance}\/ between the new
instance $X$ and the stored instance $Y$.

The most basic metric for instances with symbolic features is the {\bf
overlap metric} given in Equations~\ref{distance} and~\ref{overlap};
where $\Delta(X,Y)$ is the distance between instances $X$ and $Y$,
represented by $n$ features, $w_{i}$ is a weight for feature $i$, and
$\delta$ is the distance per feature. The $k$-{\sc nn} algorithm with
this metric, and equal weighting for all features is, for example,
implemented in {\sc ib1} \cite{Aha+91}. Usually $k$ is set to 1.
\begin{equation}
\Delta(X,Y) = \sum_{i=1}^{n}\ w_{i}\ \delta(x_{i},y_{i})
\label{distance}
\end{equation}
where:
\begin{equation}
\delta(x_{i}, y_{i}) = 0\ if\ x_{i} = y_{i},\ else\ 1
\label{overlap}
\end{equation}
We have made two additions to the original algorithm in our version of 
{\sc ib1}.  First, in the case of nearest neighbor sets larger than 
one instance ($k>1$ or ties), our version of {\sc ib1} selects the 
classification with the highest frequency in the class distribution of 
the nearest neighbor set.  Second, if a tie cannot be resolved in this 
way because of equal frequency of classes among the nearest neighbors, 
the classification is selected with the highest overall occurrence in 
the training set.

The distance metric in Equation \ref{overlap} simply counts the number
of (mis)matching feature values in both instances. In the absence of
information about feature relevance, this is a reasonable
choice. Otherwise, we can add linguistic bias to weight or select
different features \cite{Cardie96} or look at the behavior of features
in the set of examples used for training. We can compute statistics
about the relevance of features by looking at which features are good
predictors of the class labels. Information theory gives us a useful
tool for measuring feature relevance in this
way~\cite{Quinlan86,Quinlan93}.

{\bf Information gain} (IG) weighting looks at each feature in
isolation, and measures how much information it contributes to our
knowledge of the correct class label. The information gain of feature
$f$ is measured by computing the difference in uncertainty
(i.e. entropy) between the situations without and with knowledge of
the value of that feature (Equation~\ref{IGgain}).
\begin{equation}
w_{f} = \frac{H(C) -  \sum_{v \in V_{f}} P(v) H(C|v)}{si(f)}
\label{IGgain}
\end{equation}
\begin{equation}
si(f) = - \sum_{v \in V_{f}} P(v) \log_{2} P(v)
\label{splitinfo}
\end{equation}
where $C$ is the set of class labels, $V_{f}$ is the set of values for 
feature $f$, and $H(C) = - \sum_{c \in C} P(c) \log_{2} P(c)$ is the 
entropy of the class label probability distribution.  The 
probabilities are estimated from relative frequencies in the training 
set.  The normalizing factor $si(f)$ (split info) is included to avoid 
a bias in favor of features with more values.  It represents the 
amount of information needed to represent all values of the feature 
(Equation~\ref{splitinfo}).  The resulting IG values can then be used 
as weights in equation~\ref{distance}.

The possibility of automatically determining the relevance of features
implies that many different and possibly irrelevant features can be
added to the feature set. This is a very convenient methodology if
theory does not constrain the choice enough beforehand, or if we wish
to measure the importance of various information sources
experimentally. A limitation is its insensitivity to feature
redundancy; although a feature may be redundant, it may be assigned a
high information gain weight. Nevertheless, the advantages far
outweigh the limitations for our data sets, and {\sc ib1-ig}
consistently outperforms {\sc ib1}.

\subsection{C5.0}

{\sc c5.0}, a commercial version of {\sc c4.5} \cite{Quinlan93},
performs {\em top-down induction of decision trees}\/ ({\sc
tdidt}). On the basis of an instance base of examples, {\sc c5.0}
constructs a decision tree which compresses the classification
information in the instance base by exploiting differences in relative
importance of different features. Instances are stored in the tree as
paths of connected nodes ending in leaves which contain classification
information. Nodes are connected via arcs denoting feature values.
Feature information gain (Equation \ref{IGgain}) is used dynamically
in {\sc c5.0} to determine the order in which features are employed as
tests at all levels of the tree \cite{Quinlan93}.

{\sc c5.0} can be tuned by several parameters.  In our experiments, we 
chose to vary the {\em pruning confidence level}\/ (the $c$ 
parameter), and the {\em minimal number of instances represented at 
any branch of any feature-value test}\/ (the $m$ parameter).  The two 
parameters directly affect the degree of `forgetting' of individual 
instances by {\sc c5.0}:

\begin{itemize}
\item
The $c$ parameter denotes the pruning confidence level, which ranges
between 0\% and 100\%. This parameter is used in a heuristic
function that estimates the predicted number of misclassifications of
unseen instances at leaf nodes, by computing the binomial probability
(i.e, the confidence limits for the binomial distribution) of
misclassifications within the set of instances represented at that
node \cite{Quinlan93}. When the presence of a leaf node leads to a
higher predicted number of errors than when it would be absent, it is
pruned from the tree. By default, $c=25\%$; set at 100\%, no pruning
occurs. The more pruning is performed, the less information about the
individual examples is remembered in the abstracted decision tree.
\item
The $m$ parameter governs the minimum number of instances represented
by a node. By setting $m>1$, {\sc c5.0} can avoid the creation of long
paths disambiguating single-instance minorities that possibly
represent noise \cite{Quinlan93}. By default, $m=2$. With $m=1$, {\sc
c5.0} builds a path for every single instance not yet
disambiguated. Higher values of $m$ lead to an increasing amount of
abstraction and therefore to less recoverable information about
individual instances.
\end{itemize}

Moreover, we chose to set the {\em subsetting of values}\/ ($s$)
parameter at the non-default value `on'.  The $s$ parameter is a flag
determining whether different values of the same feature are grouped
on the same arc in the decision tree when they lead to identical or
highly similar subtrees. We used value grouping as a default for
reasons of computational complexity for the {\sc pos}, {\sc pp}, and
{\sc np} data sets, and because that setting yields higher
generalization accuracy for the {\sc gs} data set.

\subsection{IGTREE}

The {\sc igtree} algorithm was originally developed as a method to
compress and index case bases in memory-based learning
\cite{Daelemans+97}. It performs {\sc tdidt} in a way similar to that
of {\sc c5.0}, but with two important differences. First, it builds
{\em oblivious}\/ decision trees, i.e., feature ordering is computed
only at the root node and is kept constant during {\sc tdidt}, instead
of being recomputed at every new node. Second, {\sc igtree} does not
prune exceptional instances; it is only allowed to disregard
information redundant for the classification of the instances presented during training.

Instances are stored as paths of connected nodes and leaves in a
decision tree. Nodes are connected via arcs denoting feature
values. The global information gain of the features is used to
determine the order in which instance feature values are added as arcs
to the tree. The reasoning behind this compression is that when the
computation of information gain points to one feature clearly being
the most important in classification, search can be restricted to
matching a test instance to those memory instances that have the same
feature value as the test instance at that feature. Instead of
indexing all memory instances only once on this feature, the instance
memory can then be optimized further by examining the second most
important feature, followed by the third most important feature, etc.
A considerable compression is obtained as similar instances share
partial paths.

The tree structure is compressed even more by restricting the paths to
those input feature values that disambiguate the classification from
all other instances in the training material. The idea is that it is
not necessary to fully store an instance as a path when only a few
feature values of the instance make the instance classification
unique. This implies that feature values that do not contribute to the
disambiguation of the instance (i.e., the values of the features with
lower information gain values than the lowest information gain
value of the disambiguating features) are not stored in the tree.

Apart from compressing all training instances in the tree structure,
the {\sc igtree} algorithm also stores with each non-terminal node
information concerning the {\em most probable}\/ or {\em default}\/
classification given the path thus far, according to the bookkeeping
information maintained by the tree construction algorithm. This extra
information is essential when processing unknown test instances.
Processing an unknown input involves traversing the tree (i.e.,
matching all feature-values of the test instance with arcs in the
order of the overall feature information gain), and either retrieving
a classification when a leaf is reached (i.e., an exact match was
found), or using the default classification on the last matching
non-terminal node if an exact match fails.

In sum, in the trade-off between computation during learning and
computation during classification, the {\sc igtree} approach chooses
to invest more time in organizing the instance base than {\sc ib1-ig},
but less than {\sc c5.0}, because the order of the features needs to
be computed only once for the whole data set. 

\section{Benchmark language learning tasks}
\label{tasks}

We investigate four language learning tasks that jointly represent a
wide range of different types of tasks in the {\sc nlp} domain: (1)
grapheme-phoneme conversion (henceforth referred to as {\sc gs}), (2)
part-of-speech tagging ({\sc pos}), (3) prepositional-phrase
attachment ({\sc pp}), and (4) base noun phrase chunking ({\sc
np}). In this section, we introduce each of the four tasks, and
describe for each task the data collected and employed in our
study. First, properties of the four data sets are listed in
Table~\ref{dataset-properties}, and examples of instances for each of
the tasks are displayed in Table~\ref{instance-ex}.

\begin{table}
\begin{center}
\setlength{\tabcolsep}{0.85mm}
\begin{tabular}{|c|r|rrrrrrrrrrr|r|r|}
\hline
 & \# & \multicolumn{11}{|c|}{\scriptsize \# Values of feature} & \# & \# Data set \\
Task & Features  & 1 & 2 & 3 & 4 & 5 & 6 & 7 & 8 & 9 & 10 & 11 & Classes & instances \\
\hline
{\sc gs}  & 7 & {\scriptsize 42} & {\scriptsize 42} & {\scriptsize 42 } & {\scriptsize 41} & {\scriptsize 42} & {\scriptsize 42} & {\scriptsize 42} &    &   &   &   & 159 & 675,745 \\
{\sc pos} & 5 & {\scriptsize 170} &{\scriptsize 170} & {\scriptsize 498} & {\scriptsize 492} & {\scriptsize 480} &    &    &    &   &   &   & 169 & 1,046,152 \\
{\sc pp}  & 4 & {\scriptsize 3,474} & {\scriptsize 4,612} & {\scriptsize 68} & {\scriptsize 5,780} &     &    &    &    &   &   &   &   2 &  23,898 \\
{\sc np}  & 11 & {\scriptsize 20,231} & {\scriptsize 20,282} & {\scriptsize  20,245} & {\scriptsize 20,263} & {\scriptsize 86} & {\scriptsize 87} & {\scriptsize 86} & {\scriptsize 89} & {\scriptsize 3} & {\scriptsize 3} & {\scriptsize 3} & {\scriptsize 3} & 251,124 \\
\hline
\end{tabular}
\normalsize
\caption{Properties of the four investigated data sets of the {\sc
gs}, {\sc pos}, {\sc pp}, and {\sc np} learning tasks: numbers of
features, values per feature, classes, and instances.
\label{dataset-properties}}
\end{center}
\end{table}

\begin{table}
\setlength{\tabcolsep}{0.6mm}
\begin{center}
\begin{tabular}{|c|ccccccccccc|c|}
\hline
     & \multicolumn{11}{|c|}{Features} & \\
Task & 1 & 2 & 3 & 4 & 5 & 6 & 7 & 8 & 9 & 10 & 11 & label \\
\hline
{\sc gs}  & \_ & h & e & {\bf a} & r & t & s & & & & & 0A: \\
          & b  & o & o & {\bf k} & i & n & g & & & & & 0k \\
          & t  & i & e & {\bf s} & \_ & \_ & \_ & & & & & 0z \\
          & \_ & \_ & a & {\bf f} & a & r & \_ & & & & & 1f \\
\hline
{\sc pos} & \_ & {\sc sqso} & {\scriptsize\bf VB} & {\sc vbg} & {\sc nn} & & & & & & & {\sc vb} \\
          & {\sc nns} & {\sc bez} & {\scriptsize\bf TO/IN} & {\sc be} & {\sc vbn/vbd} & & & & & & & {\sc to} \\
          & {\sc np} & {\sc hvz} & {\scriptsize\bf VB/VBN/VBD} & {\sc rp/in} & {\sc at} & & & & & & & {\sc vbn} \\
          & \_ & \_ & {\scriptsize\bf PP3} & {\sc md} & {\sc rn} & & & & & & & {\sc pp3} \\
\hline
{\sc pp}  & is & chairman & {\bf of} & NV & & & & & & & & noun \\
          & pour & cash & {\bf into} & funds & & & & & & & & verb \\
          & asked & them & {\bf for} & views & & & & & & & & verb \\
          & caused & swings & {\bf in} & prices & & & & & & & & noun \\
\hline
{\sc np}  & definitive & agreement & {\bf between} & the & {\sc jj} & {\sc nn} & {\sc in} & {\sc dt} & I & I & I & O \\
          & when & they & {\bf need} & money & {\sc wrb} & {\sc pp} & {\sc vbp} & {\sc nn} & I & I & O & O \\
          & pose & a & {\bf new} & challenge & {\sc vb} & {\sc dt} & {\sc jj} & {\sc nn} & O & I & I & I \\
          & performance & that & {\bf would} & compare & {\sc nn} & {\sc wdt} & {\sc md} & {\sc vb} & O & B & I & O \\
\hline
\end{tabular}
\normalsize
\caption{Example of instances of the {\sc gs}, {\sc pos}, {\sc pp},
and {\sc np} learning tasks. All instances represent fixed-sized
feature-value vectors and an associated class label. Feature values
printed in bold are focus features (description in
text).\label{instance-ex}}
\end{center}
\end{table}

\subsection{GS: grapheme-phoneme conversion with stress assignment}

Converting written words to stressed phonemic transcription, i.e.,
word pronunciation, is a well-known benchmark task in machine learning
\cite{Sejnowski+87,Stanfill+86,Stanfill87,Lehnert87,Wolpert89,Shavlik+91,Dietterich+95a}. We
define the task as the conversion of fixed-sized instances
representing parts of words to a class representing the phoneme and
the stress marker of the instance's middle letter. We henceforth refer
to the task as {\sc gs}, an acronym of {\sc g}rapheme-phoneme
conversion and {\sc s}tress assignment. To generate the instances,
windowing is used \cite{Sejnowski+87}.  Table~\ref{instance-ex} (top)
displays four example instances and their
classifications. Classifications, i.e., phonemes with stress markers,
are denoted by composite labels. For example, the first instance in
Table~\ref{instance-ex}, {\sf \_hearts}, maps to class label 0A:,
denoting an elongated short `a'-sound which is not the first phoneme
of a syllable receiving primary stress. In this study, we chose a
fixed window width of seven letters, which offers sufficient context
information for adequate performance (in terms of the upper bound on
error demanded by applications in speech technology).

From {\sc celex} \cite{Baayen+93} we extracted, on the basis of the
standard word base of 77,565 words with their corresponding
transcription, a data base containing 675,745 instances. The number of
classes (i.e., all possible combinations of phonemes and stress
markers) occurring in this data base is 159.

\subsection{POS: Part-of-speech tagging of word forms in context}

Many words in a text are ambiguous with respect to their 
morphosyntactic category (part-of-speech).  Each word has a set of 
lexical possibilities, and the local context of the word can be used 
to select the most likely category from this set \cite{Church88}.  For 
example in the sentence {\em ``they can can a can''}, the word {\em 
can}\/ is tagged as modal verb, main verb and noun respectively.  We 
assume a tagger architecture that processes a sentence from the left 
to the right by classifying instances representing words in their 
contexts (as described in \namecite{Daelemans+96b}).  The word's 
already tagged left context is represented by the disambiguated 
categories of the two words to the left, the word itself and its 
ambiguous right context are represented by categories which denote 
ambiguity classes (e.g. verb-or-noun).

The data set for the part-of-speech tagging task, henceforth referred
to as the {\sc pos} task, was extracted from the LOB
corpus\footnote{The LOB corpus is available from {\sc icame}, the
International Computer Archive of Modern and Medieval English; consult
{\tt http://www.hd.uib.no/icame.html} for more information.}. The full
data set contains 1,046,152 instances. The ``lexicon'' of ambiguity
classes was constructed from the first 90\% of the corpus only, and
hence the data contains unknown words. To avoid a complicated
architecture, we treat unknown words the same as the known words,
i.e., their ambiguous category is simply ``{\sc unknown}'', and they
can only be classified on the basis of their context\footnote{In our
full POS tagger we have a separate classifier for unknown words, which
takes into account features such as suffix and prefix letters, digits,
hyphens, etc.}.

\subsection{PP: Disambiguating verb/noun attachment of prepositional phrases}

As an example of a semantic-syntactic disambiguation task we consider
a simplified version of the task of Prepositional Phrase (henceforth
{\sc pp}) attachment: the attachment of a {\tt PP} in the sequence
{\tt VP NP PP} ({\tt VP} $=$ verb phrase, {\tt NP} $=$ noun phrase,
{\tt PP} $=$ prepositional phrase). The data consists of four-tuples
of words, extracted from the Wall Street Journal Treebank
\cite{Marcus+93} by a group at {\sc ibm}
\cite{Ratnaparkhi+94}.\footnote{The data set is available from {\tt
ftp://ftp.cis.upenn.edu/pub/adwait/PPattachData/}. We would like to
thank Michael Collins for pointing this benchmark out to us.}  They
took all sentences that contained the pattern {\tt VP NP PP} and
extracted the head words from the constituents, yielding a {\tt V N1 P
N2} pattern ({\tt V} $=$ verb, {\tt N} $=$ noun, {\tt P} $=$
preposition). For each pattern they recorded whether the PP was
attached to the verb or to the noun in the treebank parse. For
example, the sentence {\em ``he eats pizza with a fork''}\/ would
yield the pattern:
\begin{quote}
\label{fork-tt}
{\tt eats, pizza, with, fork, verb.}
\end{quote}
because here the PP is an instrumental modifier of the verb. A
contrasting sentence would be {\em ``he eats pizza with anchovies''},
where the PP modifies the noun phrase {\em pizza}.
\begin{quote}
\label{anchovies-tt}
{\tt eats, pizza, with, anchovies, noun.}
\end{quote}
From the original data set, used in statistical disambiguation methods
by \namecite{Ratnaparkhi+94} and \namecite{Collins+95}, we took the
train and test set together to form a new data set of 23,898 instances.

Due to the large number of possible word combinations and the
comparatively small training set size, this data set can be considered
very sparse.  Of the 2390 test instances in the first fold of the 10
cross-validation (CV) partitioning, only 121 (5.1\%) occurred in the
training set; 619 (25.9~\%) instances had 1 mismatching word with any
instance in the training set; 1492 (62.4\%) instances had 2
mismatches; and 158 (6.6~\%) instances had 3 mismatches. Moreover, the
test set contains many words that are not present in any of the
instances in the training set.

The {\sc pp} data set is also known to be
noisy. \namecite{Ratnaparkhi+94} performed a study with three human
subjects, all experienced treebank annotators, who were given a small
random sample of the test sentences (either as four-tuples or as full
sentences), and who had to give the same binary decision. The humans,
when given the four-tuple, gave the same answer as the Treebank parse
only 88.2\% of the time, and when given the whole sentence, only
93.2\% of the time.

\subsection{NP: Base noun phrase chunking}

Phrase chunking is defined as the detection of boundaries between
phrases (e.g., noun phrases or verb phrases) in sentences.  Chunking
can be seen as a `light' form of parsing.  In {\em NP chunking},
sentences are segmented into non-recursive NP's, so called
baseNP's~\cite{Abney91}.  NP chunking can, for example, be used to
reduce the complexity of sub-sequential parsing, or to identify named
entities for information retrieval.  To perform this task, we used the
baseNP tag set as presented in~\cite{Ramshaw+95}: $I$ for inside a
baseNP, $O$ for outside a baseNP, and $B$ for the first word in a
baseNP following another baseNP. As an example, the IOB tagged
sentence: ``The/I postman/I gave/O the/I man/I a/B letter/I ./O'' will
result in the following baseNP bracketed sentence: ``[The postman]
gave [the man] [a letter].''  The data we used are based on the same
material as~\cite{Ramshaw+95} which is extracted from the Wall Street
Journal text in the parsed Penn Treebank~\cite{Marcus+93}.  Our NP
chunker consists of two stages, and in this paper we have used
instances from the second stage.  An instance (constructed for each
focus word) consists of features referring to words, POS tags, and IOB
tags (predicted by the first stage) of the focus and the two immediately
adjacent words.  The data set contains a total of 251,124 instances.

\subsection{Experimental method}
\label{expmeth}

We used 10-fold CV \cite{Weiss+91} in all experiments 
comparing classifiers (Section \ref{dectree}).  In this approach, the 
initial data set (at the level of instances) is partitioned into ten 
subsets.  Each subset is taken in turn as a test set, and the remaining 
nine combined to form the training set.  Means are reported, as well 
as standard deviation from the mean.  In the editing experiments 
(Section \ref{editing}), the first train-test partition of the 10-fold 
CV was used for comparing the effect on the test set accuracy of 
applying different editing schemes on the training set. 

Having introduced the machine learning methods and data sets that we
focus on in this paper, and the experimental method we used, the next
Section describes empirical results from a first set of experiments
aimed at getting more insight into the effect of editing exceptional
instances in memory-based learning.

\section{Editing exceptions in memory-based learning is harmful}
\label{editing}

The editing of instances from memory in memory-based learning or the 
$k$-{\sc nn} classifier \cite{Hart68,Wilson72,Devijver+80} serves two 
objectives: to minimize the number of instances in memory for reasons 
of speed or storage, and to minimize generalization error by removing 
noisy instances, prone to being responsible for generalization errors.  
Two basic types of editing, corresponding to these goals, can be found 
in the literature:

\begin{itemize}
\item
{\bf Editing superfluous regular instances}: delete instances for which
the deletion does not harm the classification accuracy of their own
class in the training set \cite{Hart68}.
\item
{\bf Editing unproductive exceptions}: deleting instances that are
incorrectly classified by their neighborhood in the training set
\cite{Wilson72}, or roughly vice-versa, deleting instances that are
bad class predictors for their neighborhood in the training set
\cite{Aha+91}.
\end{itemize}

We present experiments in which both types of editing are employed
within the {\sc ib1-ig} algorithm (Subsection~\ref{ib1ig}). The two
types of editing are performed on the basis of two criteria that
estimate the exceptionality of instances: typicality \cite{Zhang92}
and class prediction strength \cite{Salzberg90} (henceforth referred
to as {\sc cps}). Unproductive exceptions are edited by taking the
instances with the lowest typicality or {\sc cps}, and superfluous
regular instances are edited by taking the instances with the highest
typicality or {\sc cps}. Both criteria are described in
Subsection~\ref{criteria}. Experiments are performed using the {\sc
ib1-ig} implementation of the TiMBL software package\footnote{TiMBL,
which incorporates {\sc ib1-ig} and {\sc igtree} and additional
weighting metrics and search optimalizations, can be downloaded from
{\tt http://ilk.kub.nl/}.}~\cite{Daelemans+98}. We present the
results of the editing experiments in
Subsection~\ref{editing-results}.

\subsection{Two editing criteria}
\label{criteria}

We investigate two methods for estimating the (degree of)
exceptionality of instance types: typicality and class prediction
strength ({\sc cps}).

\subsubsection{Typicality}

In its common meaning, ``typicality'' denotes roughly the opposite of
exceptionality; atypicality can be said to be a synonym of
exceptionality.  We adopt a definition from \cite{Zhang92}, who
proposes a typicality function. Zhang computes typicalities of
instance types by taking the notions of {\em intra-concept
similarity}\/ and {\em inter-concept similarity}\/ \cite{Rosch+75}
into account. First, Zhang introduces a distance function which
extends Equation~\ref{distance}; it normalizes the distance between
two instances $X$ and $Y$ by dividing the summed squared distance by
$n$, the number of features. The normalized distance function used by
Zhang is given in Equation~\ref{zhangdist}.
\begin{equation}
\Delta(X,Y)=\sqrt{\frac{1}{n} \sum_{i=1}^{n} (\delta(x_{i},y_{i}))^{2} }
\label{zhangdist}
\end{equation}
The intra-concept similarity of instance $X$ with classification $C$
is its similarity (i.e., $1-$distance) with all instances in the data
set with the same classification $C$: this subset is referred to as
$X$'s {\em family}, $Fam(X)$. Equation~\ref{intra} gives the
intra-concept similarity function $Intra(X)$ ($|Fam(X)|$ being the
number of instances in $X$'s family, and $Fam(X)_{i}$ the $i$th
instance in that family).
\begin{equation}
Intra(X)=\frac{1}{|Fam(X)|} \sum_{i=1}^{|Fam(X)|} 1.0 - \Delta(X,Fam(X)_{i}) 
\label{intra}
\end{equation}
All remaining instances belong to the subset of unrelated instances,
$Unr(X)$. The inter-concept similarity of an instance $X$, $Inter(X)$,
is given in Equation~\ref{inter} (with $|Unr(X)|$ being the number of
instances unrelated to $X$, and $Unr(X)_{i}$ the $i$th instance in
that subset).
\begin{equation}
Inter(X)=\frac{1}{|Unr(X)|} \sum_{i=1}^{|Unr(X)|} 1.0 - \Delta(X,Unr(X)_{i}) 
\label{inter}
\end{equation}
The typicality of an instance $X$, $Typ(X)$, is $X$'s
intra-concept similarity divided by $X$'s inter-concept similarity, as given
in Equation~\ref{typicalityfunc}.
\begin{equation}
Typ(X)=\frac{Intra(X)}{Inter(X)}
\label{typicalityfunc}
\end{equation}
An instance type is typical when its intra-concept similarity is
larger than its inter-concept similarity, which results in a
typicality larger than 1. An instance type is atypical when its
intra-concept similarity is smaller than its inter-concept similarity,
which results in a typicality between 0 and 1. Around typicality value
1, instances cannot be sensibly called typical or atypical;
\namecite{Zhang92} refers to such instances as {\em boundary}\/ instances.

We adopt typicality as an editing criterion here, and use it for
editing instances with low typicality as well as instances with high
typicality. Low-typical instances can be seen as exceptions, or bad
representatives of their own class and could therefore be pruned from
memory, as one can argue that they cannot support productive
generalizations. This approach has been advocated by
\namecite{Ting94b} as a method to achieve significant improvements in
some domains.  Editing atypical instances would, in this line of
reasoning, not be harmful to generalization, and chances are that
generalization would even improve under certain conditions
\cite{Aha+91}.  High-typical instances, on the other hand, may be good
predictors for their own class, but there may be enough of them in
memory, so that a few may also be edited without harmful effects to
generalization.

Table~\ref{typ-examples} provides examples of low-typical (for each
task, the top three) and high-typical (bottom three) instances of all
four tasks.  The {\sc gs} examples show that loan words such as {\sf
czech} introduce peculiar spelling-pronunciation relations;
particularly foreign spellings turn out to be low-typical.
High-typical instances are parts of words of which the focus letter is
always pronounced the same way.  Low-typical {\sc pos} instances tend
to involve inconsistent or noisy associations between an unambiguous
word class of the focus word and a different word class as
classification: such inconsistencies can be largely attributed to
corpus annotation errors.  Focus tags of high-typical {\sc pos}
instances are already unambiguous.  The examples of low-typical {\sc
pp} instances represent minority exceptions or noisy instances in
which it is questionable whether the chosen classification is right
(recall that human annotators agree only on 88\% of the instances in
the data set, cf.  Subsection~\ref{tasks}), while the high-typical
{\sc pp} examples have the preposition `of' in focus position, which
typically attaches to the noun.  Low-typical {\sc np} instances seem
to be partly noisy, and otherwise difficult to interpret.
High-typical {\sc np} instances are clear-cut cases in which a noun
occurring between a determiner and a finite verb is correctly
classified as being inside an {\tt NP}.

\begin{table}
\begin{center}
\begin{tabular}{|lcr|}
\hline
\multicolumn{3}{|c|}{{\sc gs}} \\
feature values & class & typicality \\
\hline
{\sf u r e a u c r}    & 0@U  &  0.43 \\
{\sf f r e u d i a}    & 0OI  &  0.44 \\
{\sf \_ \_ c z e c h}  & 0-   &  0.54 \\
\hline
{\sf b j e c t i o}    & 0kS  & 10.57 \\
{\sf l k - o v e r}    & 2@U  & 10.39 \\
{\sf e y - j a c k}    & 2\_  &  9.41 \\
\hline
\multicolumn{3}{|c|}{{\sc pos}} \\
feature values & class & typicality \\
\hline
{\small\sc sxm sqsc cc to/in vb}          & {\sc fw}    &    0.05 \\
{\small\sc cd nnu nn bo aa}               & {\sc aq}    &    0.07 \\
{\small\sc pp3os do cc vb pp3as}          & {\sc cs}    &    0.08 \\
\hline
{\small\sc cs3 cs4 pp1as nn/jjb/in pp3os} & {\sc pp1as} & 3531.53 \\
{\small\sc cs1 cs2 cd nnu1/in nnu2}       & {\sc cd}    & 2887.29 \\
{\small\sc nn2 in2 cd nnu/zz in/cc}       & {\sc cd}    & 2526.98 \\
\hline
\multicolumn{3}{|c|}{{\sc pp}} \\
feature values & class & typicality \\
\hline
{\small accuses Motorola of turnabout} & verb & 0.01 \\
{\small cleanse Germany of muck}       & verb & 0.01 \\
{\small directs flow through systems}  & noun & 0.02 \\
\hline
{\small excluding categories of food}  & noun & 94.52 \\
{\small underscoring lack of stress}   & noun & 94.52 \\
{\small calls frenzy of legislating}   & noun & 94.53 \\
\hline
\multicolumn{3}{|c|}{{\sc np}} \\
feature values & class & typicality \\
\hline
{\small generally a bit safer {\sc rb dt nn jjr} O O O}           & O & 0.27 \\
{\small `` No matter how `` {\sc dt nn wrb} O O O}                & O & 0.27 \\
{\small I know that voluntarily {\sc pp vbp in rb} O O B}         & I & 0.27 \\
\hline
{\small that the legislator wins {\sc in dt nn vbz} O B B}        & I & 6.93 \\
{\small that the bank supports {\sc in dt nn vbz} O B B}          & I & 6.94 \\
{\small that the company hopes {\sc in dt nn vbz} O B B}          & I & 6.97 \\
\hline
\end{tabular}
\caption{Examples of low-typical (top three) and high-typical (bottom
three) instances of the {\sc gs}, {\sc pos}, {\sc pp}, and {\sc np}
learning tasks. For each instance its typicality value is
given.\label{typ-examples}}
\end{center}
\end{table}

\subsubsection{Class-prediction strength}

A second estimate of exceptionality is to measure how well an instance
type predicts the class of all other instance types within the
training set. Several functions for computing class-prediction
strength have been proposed, e.g., as a criterion for removing
instances in memory-based ($k$-{\sc nn}) learning algorithms, such as
{\sc ib3} \cite{Aha+91} (cf. earlier work on edited $k$-{\sc nn}
\cite{Hart68,Wilson72,Devijver+80,Voisin+87}); or for weighting
instances in the {\sc Each} algorithm \cite{Salzberg90}. We use the
class-prediction strength function as proposed by
\namecite{Salzberg90}. This is the ratio of the number of times the
instance type is a nearest neighbor of another instance with the same
class and the number of times that the instance type is the nearest
neighbor of another instance type regardless of the class. An instance
type with class-prediction strength 1.0 is a perfect predictor of its
own class; a class-prediction strength of 0.0 indicates that the
instance type is a bad predictor of classes of other instances,
presumably indicating that the instance type is exceptional. Even more
than with typicality, one might argue that bad class predictors can be
edited from the instance base. Likewise, one could also argue that
instances with a maximal {\sc cps} could be edited to some degree too
without harming generalization: strong class predictors may be
abundant and some may be safely forgotten since other instance types
may be strong enough to support the class predictions of the edited
instance type.

In Table~\ref{strength-examples}, examples from the four tasks of
instances with low (top three) and high (bottom three) {\sc cps} are
displayed. Many instances with low {\sc cps} are {\em minority
ambiguities}. For instance, the {\sc gs} examples represent instances
which are completely ambiguous and of which the classification is the
minority. For example, there are more words beginning with {\sf algo}
that have primary stress (class `1ae') than secondary stress (class
`2ae'), which makes the instance `\_\_\_algo 2ae' a minority
ambiguity.

\begin{table}
\begin{center}
\begin{tabular}{|lcr|}
\hline
\multicolumn{3}{|c|}{{\sc gs}} \\
feature values & class & cps \\
\hline
{\sf \_ \_ \_ a l g o} & 2ae & 0.00 \\
{\sf c k - b e n c}    & 1b  & 0.00 \\
{\sf e r b y \_ \_ \_} & 0aI & 0.00 \\
\hline
{\sf \_ \_ \_ w e e k} & 1w  & 1.00 \\
{\sf a i n d e r s}    & 0d  & 1.00 \\
{\sf e r a c t e d}    & 0k  & 1.00 \\ 
\hline
\multicolumn{3}{|c|}{{\sc pos}} \\
feature values & class & cps \\
\hline
{\small\sc scom npt in np np/nn}  & {\sc in}   & 0.00 \\
{\small\sc == == npt np genm/bez} & {\sc nn}   & 0.00 \\
{\small\sc ati nns vbn/vbd in np} & {\sc vbd}  & 0.00 \\
\hline
{\small\sc sqso wrb xnot vb ati}  & {\sc xnot} & 1.00 \\
{\small\sc ber cd nns in nn}      & {\sc nns}  & 1.00 \\
{\small\sc at jnp nn vbz in}      & {\sc nn}   & 1.00 \\
\hline
\multicolumn{3}{|c|}{{\sc pp}} \\
feature values & class & cps \\
\hline
{\small allowed access notwithstanding designations} & verb & 0.00 \\
{\small had yield during week}                       & noun & 0.00 \\
{\small make commodity of luxury}                    & verb & 0.02 \\
\hline
{\small is one of strategy}                          & noun & 0.99 \\
{\small is one of restructuring}                     & noun & 0.99 \\
{\small is one of program}                           & noun & 0.99 \\
\hline
\multicolumn{3}{|c|}{{\sc np}} \\
feature values & class & cps \\
\hline
{\small of KLM Royal Dutch {\sc in np np np} I I O}          & I & 0.00 \\
{\small in ethics charges against {\sc in nns nns in} O I O} & I & 0.00 \\
{\small assets . The axiom {\sc nns stop dt nn} I O I}       & I & 0.00 \\
\hline
{\small I drink to your {\sc pp vbp to pp} I O I}            & I & 1.00 \\
{\small share price could zoom {\sc nn nn md vb} I I O}      & O & 1.00 \\
{\small work force as well {\sc nn nn rb rb} O I I}          & O & 1.00 \\
\hline
\end{tabular}
\caption{Examples of instances with low class prediction strength (top
three) and high class prediction strength (bottom three) of the {\sc
gs}, {\sc pos}, {\sc pp}, and {\sc np} tasks.  For each instance its
class prediction strength (cps) value is given.\label{strength-examples}}
\end{center}
\end{table}

To test the utility of these measures as criteria for justifying
forgetting of specific training instances, we performed a series of
experiments in which {\sc ib1-ig} is applied to the four data sets,
systematically edited according to each of four tested criteria. We
performed the editing experiments on the first fold of the 10-fold CV
partitioning of the four data sets.  For each editing criterion (i.e.,
low and high typicality, and low and high {\sc cps}), we created eight
edited instance bases by removing 1\%, 2\%, 5\%, 10\%, 20\%, 30\%,
40\%, and 50\% of the instance tokens (rounded off so as to remove a
whole number of instance types) according to the criterion from a
single training set (the training set of the first 10-fold CV
partition). {\sc ib1-ig} was then trained on each of the edited
training sets, and tested on the original unedited test set (of the
first 10-fold CV partition). 

\begin{figure}
\centerline{
        \epsfxsize=0.5\textwidth
        \epsfbox{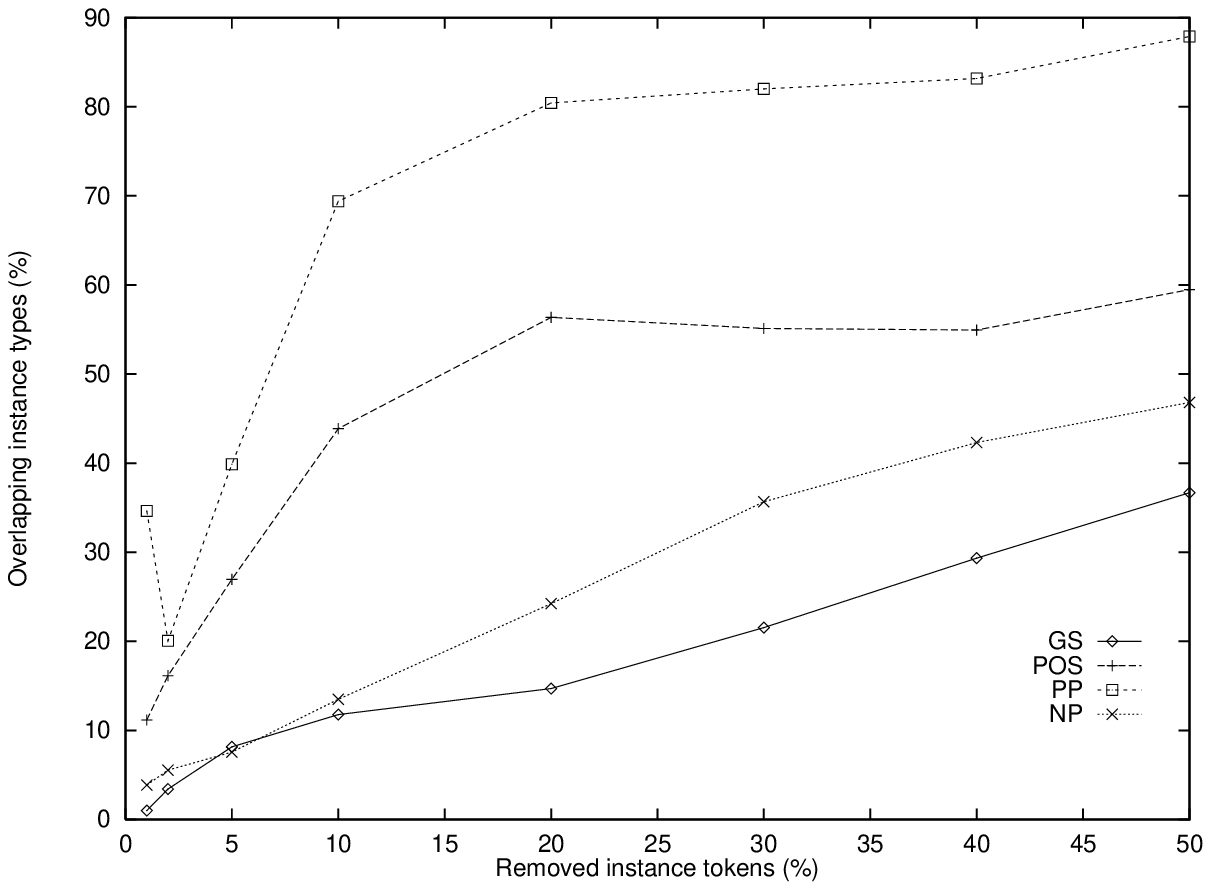}
        \epsfxsize=0.5\textwidth
        \epsfbox{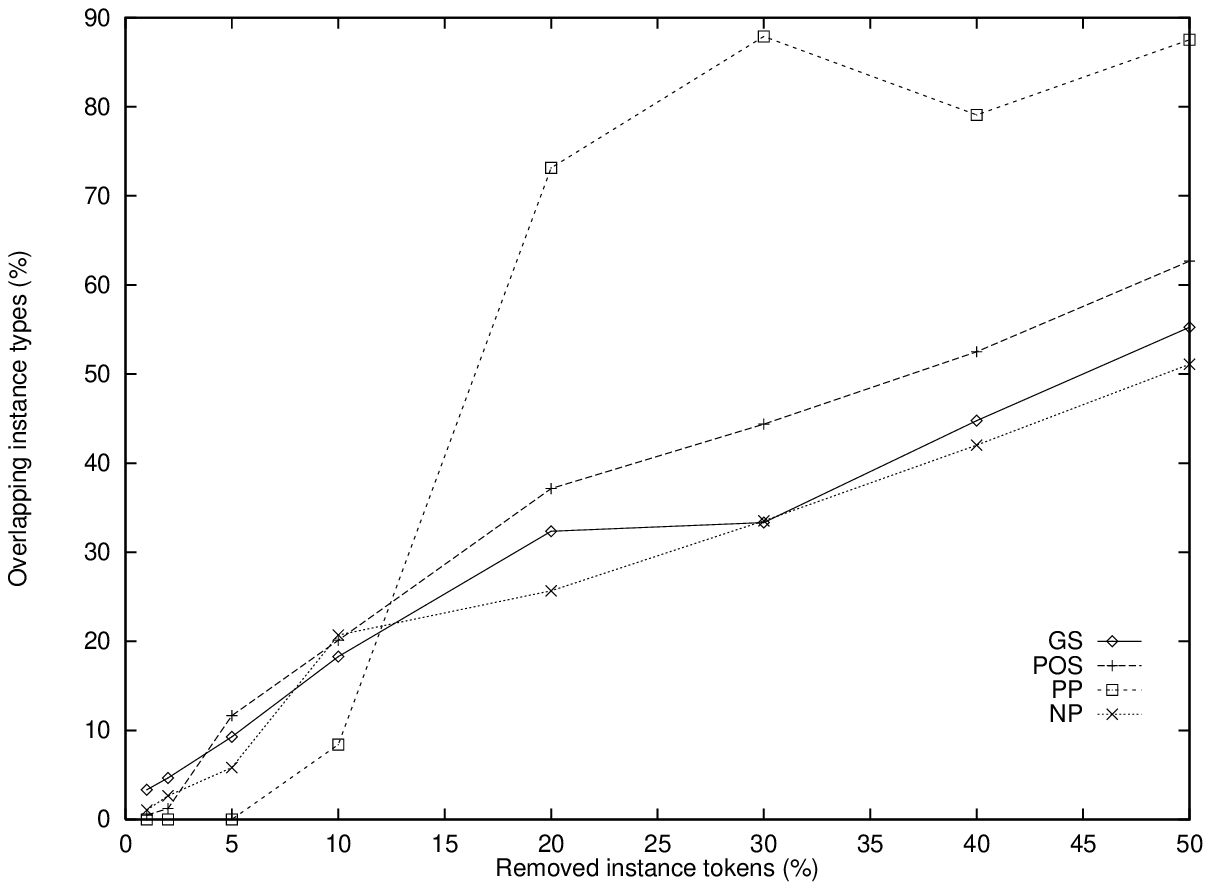}
}
\caption{The percentage of instance types that are edited by both the
typicality and the class prediction strength criterion. The left part
of the figure shows the results for editing exceptional instances, the
right part shows the results for editing regular instances.}
\label{edit-overlap}
\end{figure}

To measure to what degree the two criteria are indeed {\em different}\/
measures of exceptionality, the percentage of overlap between the
removed types was measured for each data set. As can be seen in
Figure~\ref{edit-overlap}, the two measures mostly have fairly little
overlap, certainly for editing below 10\%. The reason for this is that
typicality is based on global properties of the data set, whereas class
prediction strength is based only on the local neighborhood of each
instance. Only for the PP attachment and POS tagging tasks do the sets
of edited exceptional instances overlap up to 70\% when editing 10\%.

\subsection{Editing exceptions: Results}
\label{editing-results}

\begin{figure}
\centerline{
        \epsfxsize=0.5\textwidth
        \epsfbox{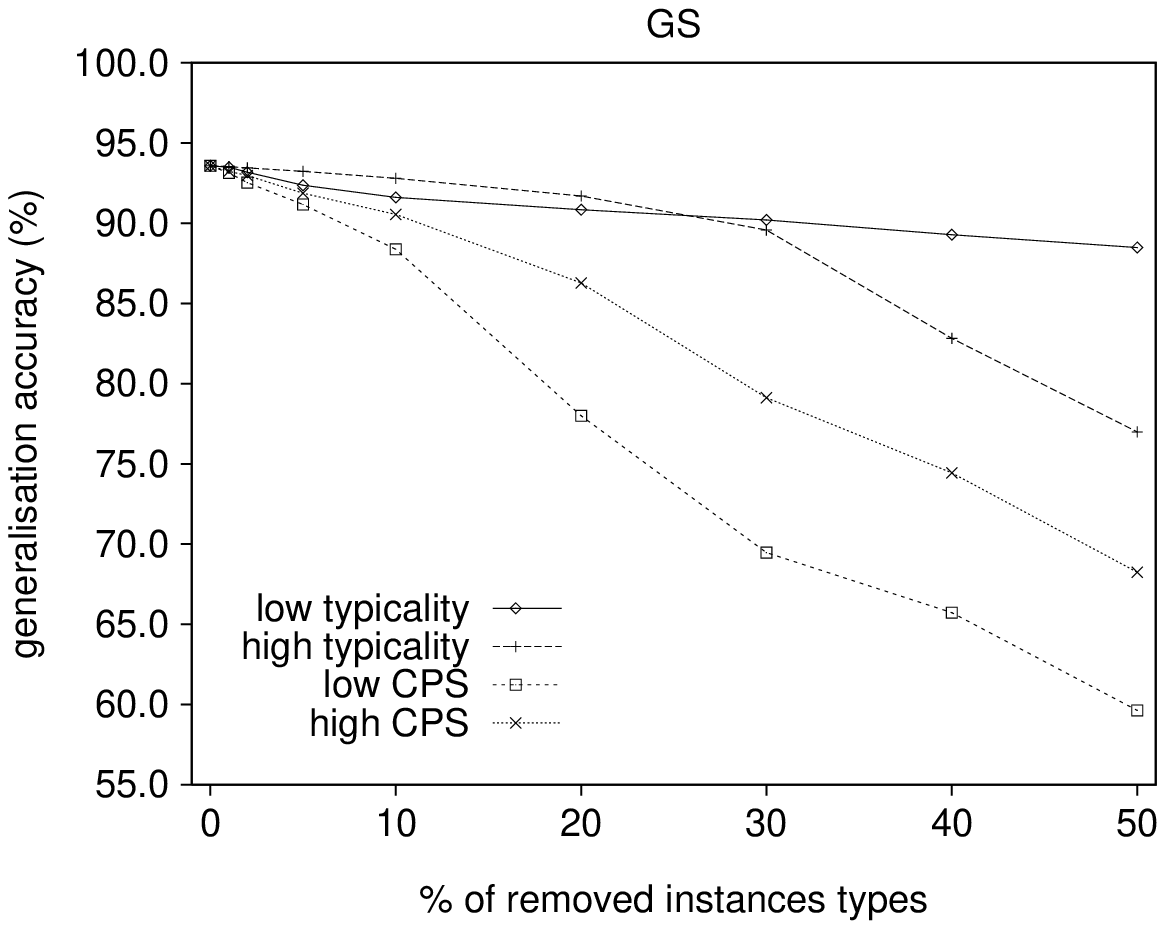}
        \epsfxsize=0.5\textwidth
        \epsfbox{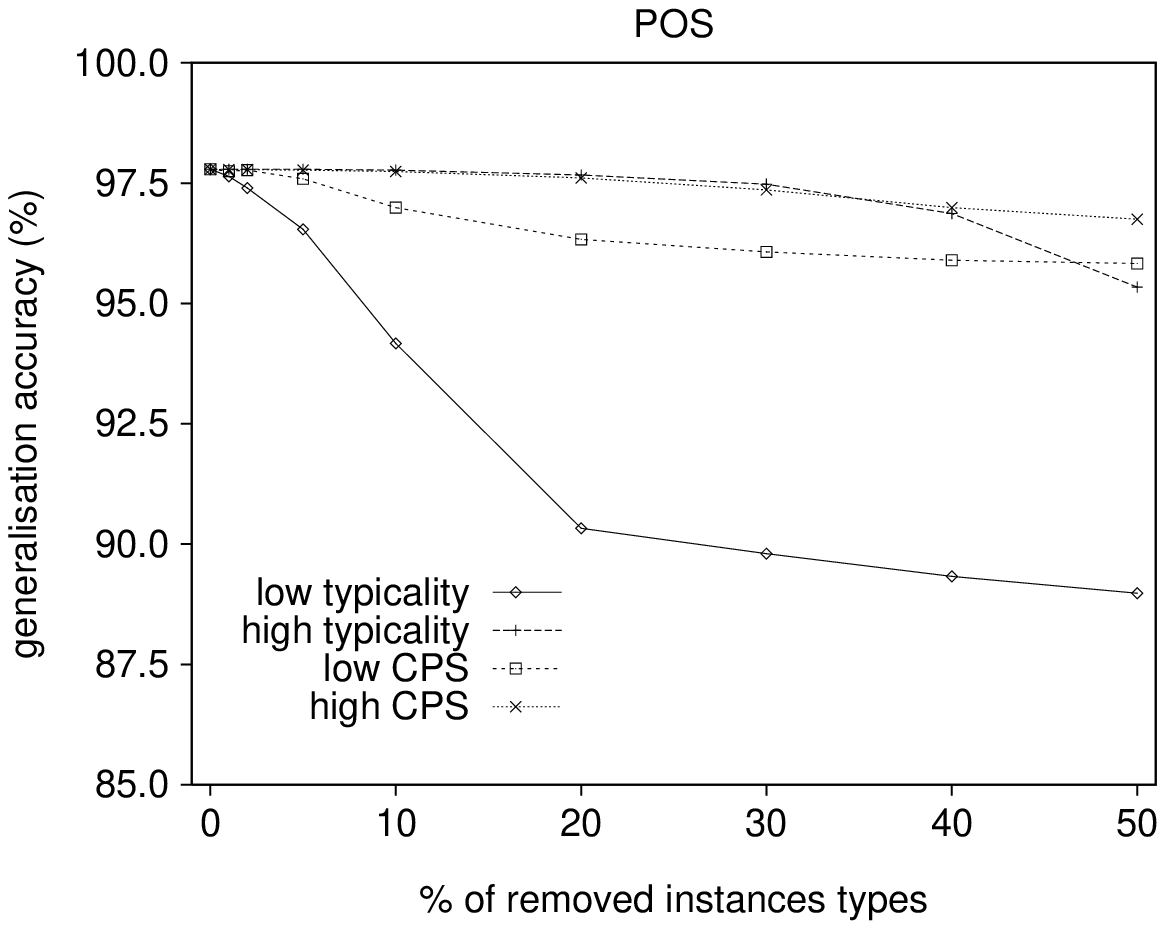}
}
\ \\
\centerline{
        \epsfxsize=0.5\textwidth
        \epsfbox{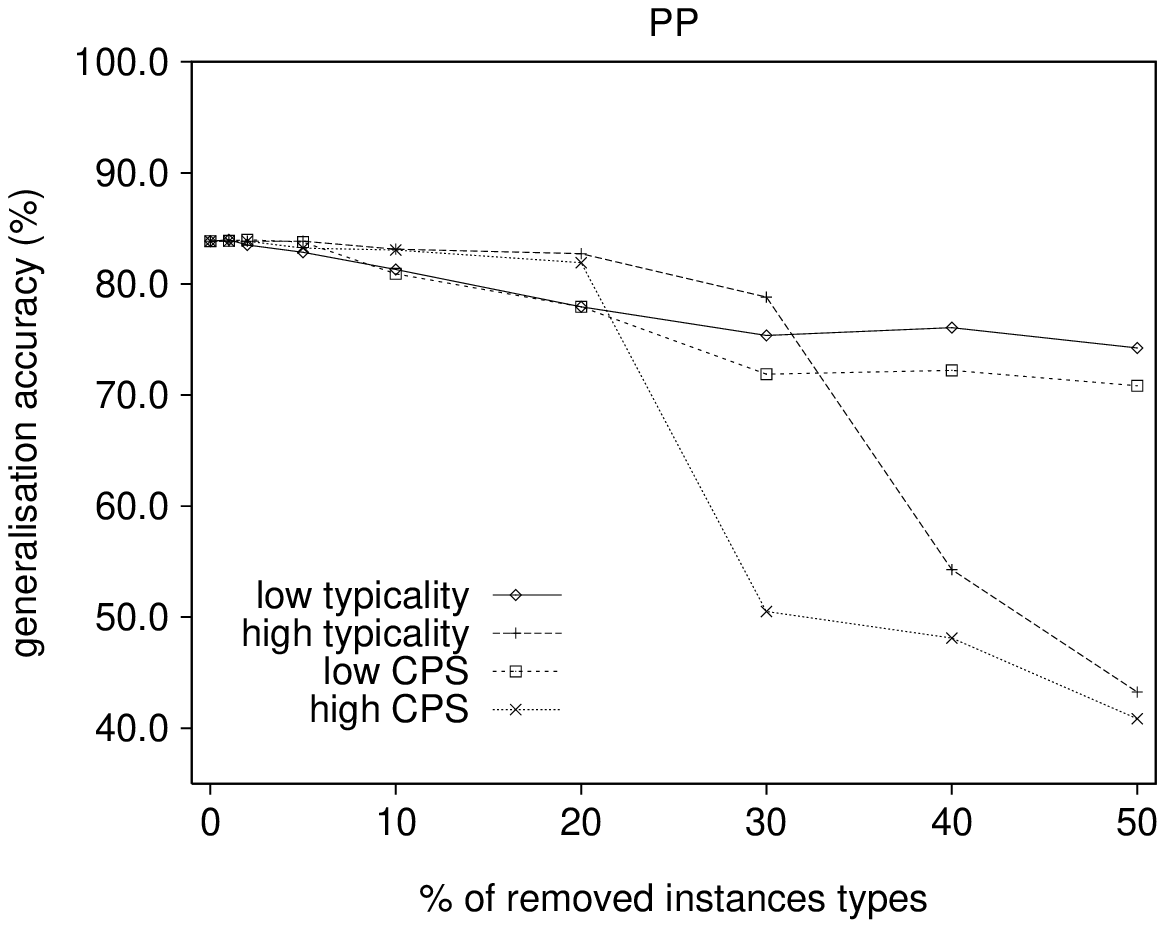}
        \epsfxsize=0.5\textwidth
        \epsfbox{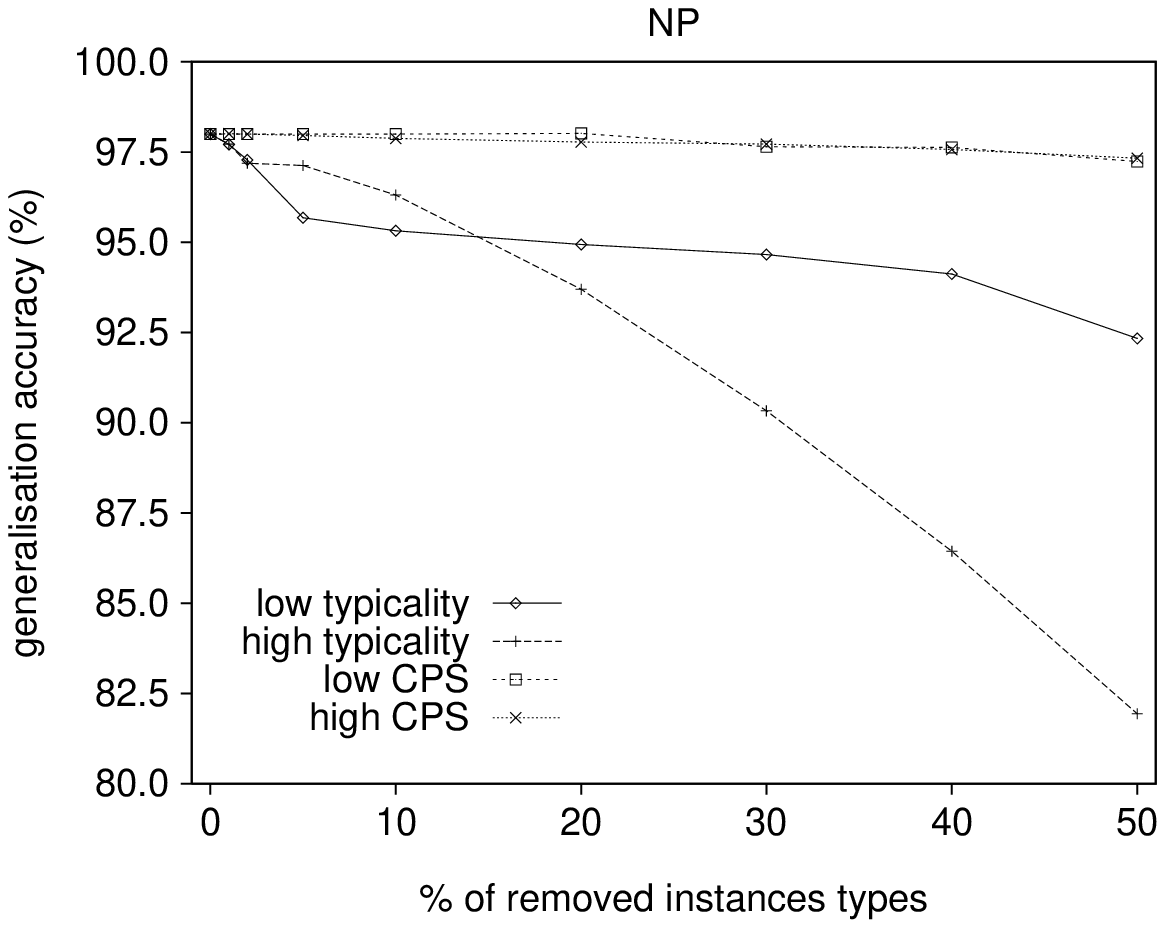}
}
\caption{Generalization accuracies (in terms of \% of correctly
classified test instances) of {\sc ib1-ig} on the four tasks with
increasing percentages of edited instance tokens, according to the four
tested editing criteria.}
\label{ib1ig-editing-perf}
\end{figure}

The general trend we observe in the results obtained with the editing
experiments is that editing on the basis of typicality and
class-prediction strength, whether low or high, is not beneficial, and
is ultimately harmful to generalization accuracy.  More specifically,
we observe a trend that editing instance types with high typicality or
high {\sc cps} is less harmful than editing instance types with low
typicality or low class prediction strength -- again, with some
exceptions.  The results are summarized in
Figure~\ref{ib1ig-editing-perf}.  The results show that in any case
for our data sets, editing serves neither of its original goals.  If
the goal is a decrease of speed and memory requirements, editing
criteria should allow editing of 50\% or more without a serious
decrease in generalization accuracy.  Instead, we see disastrous
effects on generalization accuracy at much lower editing rates,
sometimes even at 1\%.  When the goal is {\em improving}\/
generalization accuracy by removing noise, the focus of the editing
experiments in this paper, none of the studied criteria turns out to
be useful. 

To compute the statistical significance of the effect of editing, the
output for each criterion was compared to the correct classification
and the output of the unedited classifier. The resulting
cross-tabulation of hits and misses was subjected to McNemar's $\chi^2$
test~\cite{DietterichIp}. Differences with $p<0.05$ are reported as
significant.

A detailed look at the results per data set shows the following
results.  Editing experiments on the {\sc gs} task (top left of
Figure~\ref{ib1ig-editing-perf}) show significant decreases in
generalization accuracy with all editing criteria and all amounts
(even 1\% is harmful); editing on the basis of low and high {\sc cps}
is particularly harmful, and all criteria except low typicality show a
dramatic drop in accuracy at high levels of editing.

The editing results on the {\sc pos} task (top right of Figure
~\ref{ib1ig-editing-perf}) indicate that editing on the basis of
either low typicality or low class prediction strength leads to
significant decreases in generalization accuracy even with the
smallest amount (1\%) of edited instance types. Editing on the basis
of high typicality and high {\sc cps} can be performed up to 10\% and
5\% respectively without significant performance loss. For this data
set, the drop in performance is radical only for low typicality.

Editing on the {\sc pp} task (bottom left of Figure
~\ref{ib1ig-editing-perf}) results in significant decreases of
generalization accuracy with respectively 5\% and 10\% of edited
instance tokens of low typicality and low {\sc cps}. Editing with high
typicality and high {\sc cps} can be performed up to 20\% and 10\%
repectively, without significant performance loss, but accuracies drop
dramatically when 30\% or more of high-typical or high-{\sc cps}
instance types are edited.

Finally, editing on the {\sc np} data (bottom right of Figure
~\ref{ib1ig-editing-perf}) can be done without significant
generalization accuracy loss with either the low or the high {\sc cps}
criterion, up to respectively 30\% and 10\%. Editing with low or high
typicality, however, is harmful to generalization immediately from
editing 1\% of the instance tokens.

In sum, the experiments with editing on the basis of criteria
estimating the exceptionality of instances show that forgetting of
exceptional instances in memory-based learning while safeguarding
generalization accuracy can only be performed to a very limited degree
by (i) replacing instance tokens by instance types with frequency
information (which is trivial and is done by default in {\sc ib1-ig}),
and (ii) removing small amounts of minority ambiguities with low (0.0)
{\sc cps}.  None of the editing criteria studied is able to reliably
filter out noisy instances.  It seems that for the linguistic tasks we
study, methods filtering out noise tend to also intercept at least
some (small families of) productive instances.  Our experiments show
that there is little reason to believe that such editing will lead to
accuracy improvement.  When looking at editing from the perspective of
reducing storage requirements, we find that the amount of editing
possible without a significant decrease in generalization accuracy is
limited to around 10\%. Whichever perspective is taken, there does not
seem to be a clear pattern across the data sets favoring either the
typicality or class prediction strength criterion, which is somewhat
surprising given their different basis (i.e., as a measure of global or
local exceptionality).

\section{Forgetting by decision-tree learning can be harmful in 
language learning}
\label{dectree}

Another way to study the influence of exceptional instances on
generalization accuracy is to compare {\sc ib1-ig}, without editing,
to inductive algorithms that abstract from exceptional instances by means
of pruning or other devices.  {\sc c5.0} and {\sc igtree}, introduced
in Section \ref{methods} are decision tree learning methods that
abstract in various ways from exceptional instances.  We compared the
three algorithms for all data sets using 10-fold CV. In this Section,
we will discuss the results of this comparison, and the influence of
some pruning parameters of {\sc c5.0} on generalization accuracy.

\subsection{Results}

Ordered on a continuum representing how exceptional instances are
handled, {\sc ib1-ig} is at one end, keeping all training data, and
{\sc c5.0} with default settings ($c=25$, $m=2$, value grouping on) is
at the other end, making abstraction from exceptional (noisy)
instances by pruning, constructing features (by grouping subsets of
values of a feature), and enforcing a minimal number of instances at
each node.  In between is {\sc igtree}, which collapses instances that
have the same class and the same values for the most relevant features
into one node.

\begin{table}
\begin{center}
\begin{tabular}{|l|rr|rr|rr|}
\hline
 & \multicolumn{6}{|c|}{Generalization accuracy} \\
 & \multicolumn{2}{|c|}{\sc ib1-ig} & 
   \multicolumn{2}{|c|}{\sc igtree} &
   \multicolumn{2}{|c|}{\sc c5.0} \\
Task & \% & $\pm$ & \% & $\pm$ & \% & $\pm$ \\
\hline
{\sc gs}  & 93.45 & 0.15 & 93.09 & 0.15 & 92.48 & 0.14 \\
{\sc pos} & 97.94 & 0.05 & 97.75 & 0.03 & 97.97 & 0.04 \\
{\sc pp}  & 83.48 & 1.16 & 78.28 & 1.79 & 80.89 & 1.01 \\
{\sc np}  & 98.07 & 0.05 & 97.28 & 0.08 &  ---  &  --- \\
\hline
\end{tabular}
\caption{Generalization accuracies (in terms of percentages of 
correctly classified test instances) on the {\sc gs}, {\sc pos}, {\sc 
pp}, and {\sc np} tasks, by {\sc ib1-ig}, {\sc igtree}, and {\sc c5.0} 
with parameter setting $c=25$ and $m=2$ (default setting).}
\label{overall-perf}
\end{center}
\end{table}

Table~\ref{overall-perf} displays the generalization accuracies, 
measured in percentages of correctly classified test instances, for 
{\sc ib1-ig}, {\sc igtree}, and {\sc c5.0} on the four tasks.  We were 
unfortunately unable to finish the {\sc c5.0} experiment on the {\sc 
np} data set for memory reasons (running on a SUN Sparc 5 with 160 Mb internal memory and 386 Mb swap space).  The 
statistical significance of the differences between the algorithms is 
summarized in Table~\ref{signif-perf}.  We performed a one-tailed 
paired t-test between the results of the 10 CV runs.


As the results in these Tables show, {\sc ib1-ig} has significantly
better generalization accuracy than {\sc igtree} for all data sets.  In
two of the three data sets where the comparison is feasible, {\sc
ib1-ig} performs significantly better than {\sc c5.0}.  For the {\sc
pos} data set, {\sc c5.0} outperforms {\sc ib1-ig} with a small but
statistically significant difference.

\begin{table}
\begin{center}
\setlength{\tabcolsep}{0.8mm}
\begin{tabular}{|cc|c|c|c|c|}
\hline
Algorithm 1 & Algorithm 2     & {\sc gs} & {\sc pos} & {\sc pp} & {\sc np}\\
\hline
{\sc ib1-ig} &  {\sc c5.0} & $>$ ($p<10^{-6}$) & $<$ ($p=4 \times 10^{-4}$) & $>$ ($p=2 \times 10^{-4}$) & {\sc na}\\
\hline
{\sc ib1-ig} &  {\sc igtree}  & $>$ ($p<10^{-6}$) & $>$ ($p<10^{-6}$) & $>$ ($p<10^{-6}$) & $>$ ($p<10^{-6}$) \\
\hline
{\sc igtree} & {\sc c5.0}  & $>$ ($p<10^{-6}$) & $<$ ($p<10^{-6}$) & $<$ ($p=10^{-4}$) & {\sc na}\\
\hline
\end{tabular}
\caption{Significance of the differences between the generalization
performances of {\sc ib1-ig}, {\sc c5.0opt}, {\sc c5.0def}, and {\sc
igtree}, for the four tasks. A one-tailed paired t-test ($df = 9$) was
performed, to see whether the generalization accuracy of the algorithm
to the left is better than that of the algorithm to the right
(indicated by a greater than ``$>$'' sign), or the other way around
(less than sign ``$<$'').\label{signif-perf}}
\end{center}
\end{table}

\subsubsection{Abstraction in C5.0}

We performed additional experiments with {\sc c5.0} with increasing
values for the $c$ and $m$ parameters, to gain more insight into the
effect of explicitly forgetting feature-value information through
pruning ($c$) or blocking the disambiguation of small amounts of
instances ($m$).  The following space of parameters was explored for
each data set on the first fold of the 10 CV partitioning.

	\begin{enumerate}	
	\item
	$m=1$ and $c={100,75,50,40,35,30,25,20,15,10,5,2,1}$ to visualize the
	gradual increase of pruning, and
	\item
	$c=100$ and $m={1,2,3,4,5,6,8,10,15,20,30,50}$ to visualize the
	gradual decrease in the level of instance granularity at feature
        tests.
	\end{enumerate}

\begin{figure}
\centerline{
        \epsfxsize=0.5\textwidth
        \epsfbox{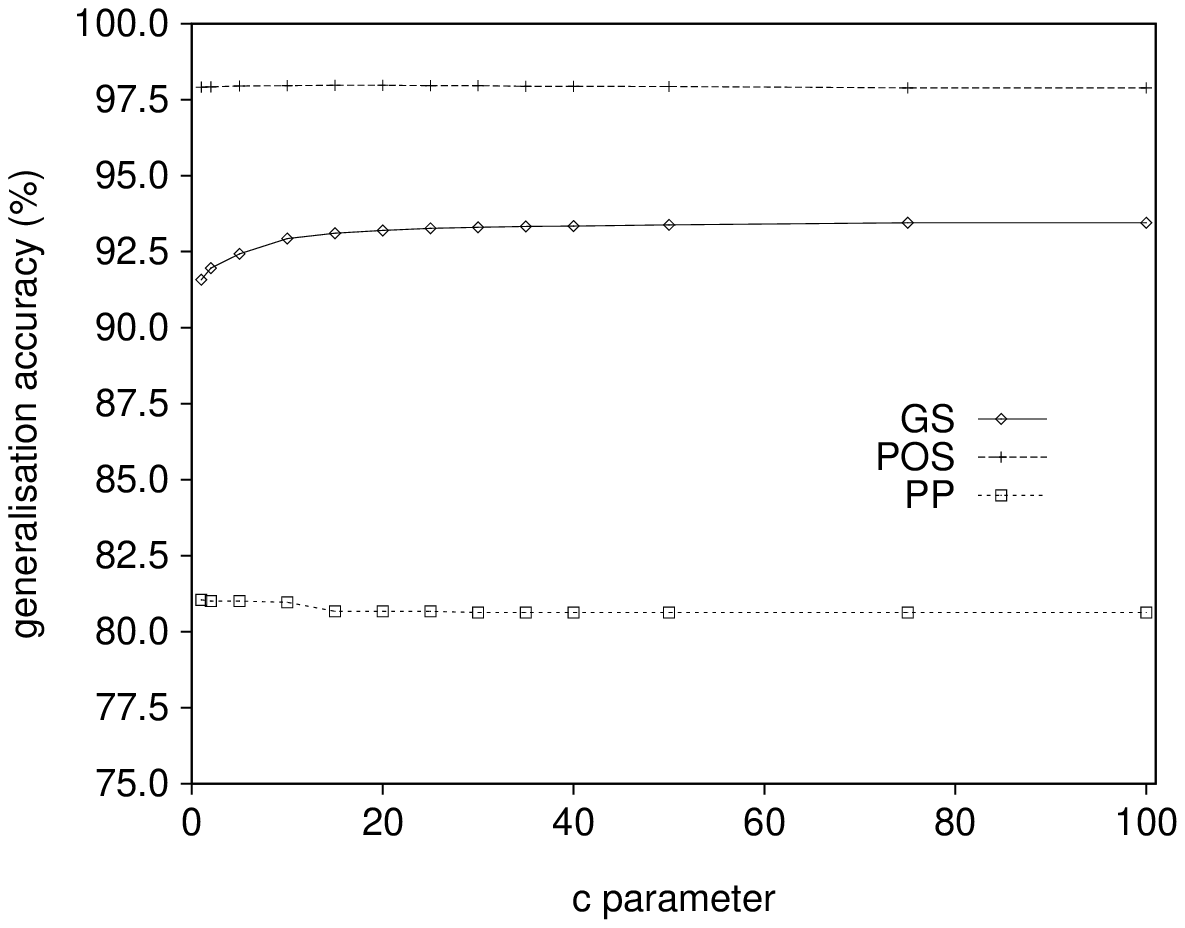}
        \epsfxsize=0.5\textwidth
        \epsfbox{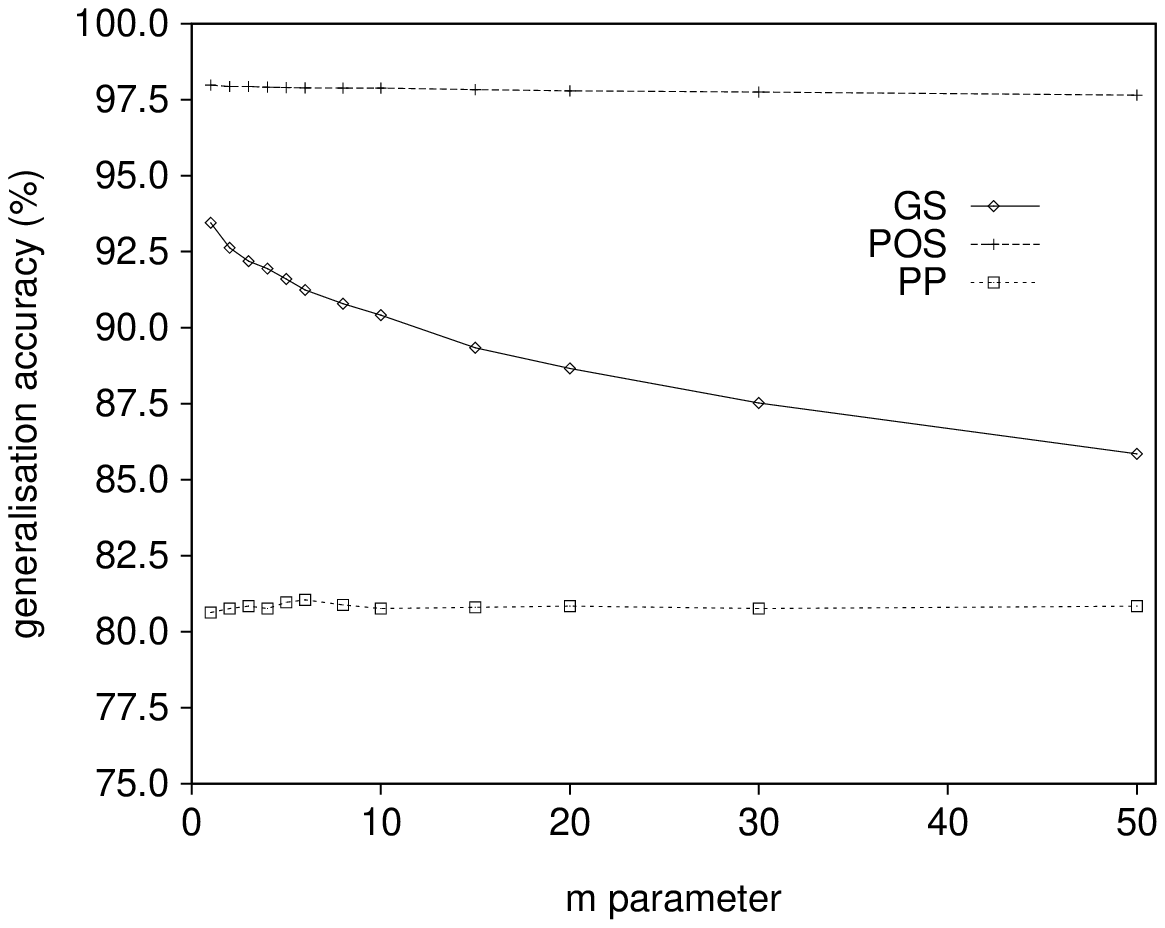}
}
\caption{Generalization accuracies (in terms of \% of correctly
classified test instances) of {\sc c5.0} with increasing $c$ parameter
(left) and increasing $m$ parameter (right), for the {\sc gs}, {\sc
pos}, and {\sc pp} tasks.}
\label{c50-parameter-perf}
\end{figure}

\begin{figure}
\centerline{
        \epsfxsize=0.5\textwidth
        \epsfbox{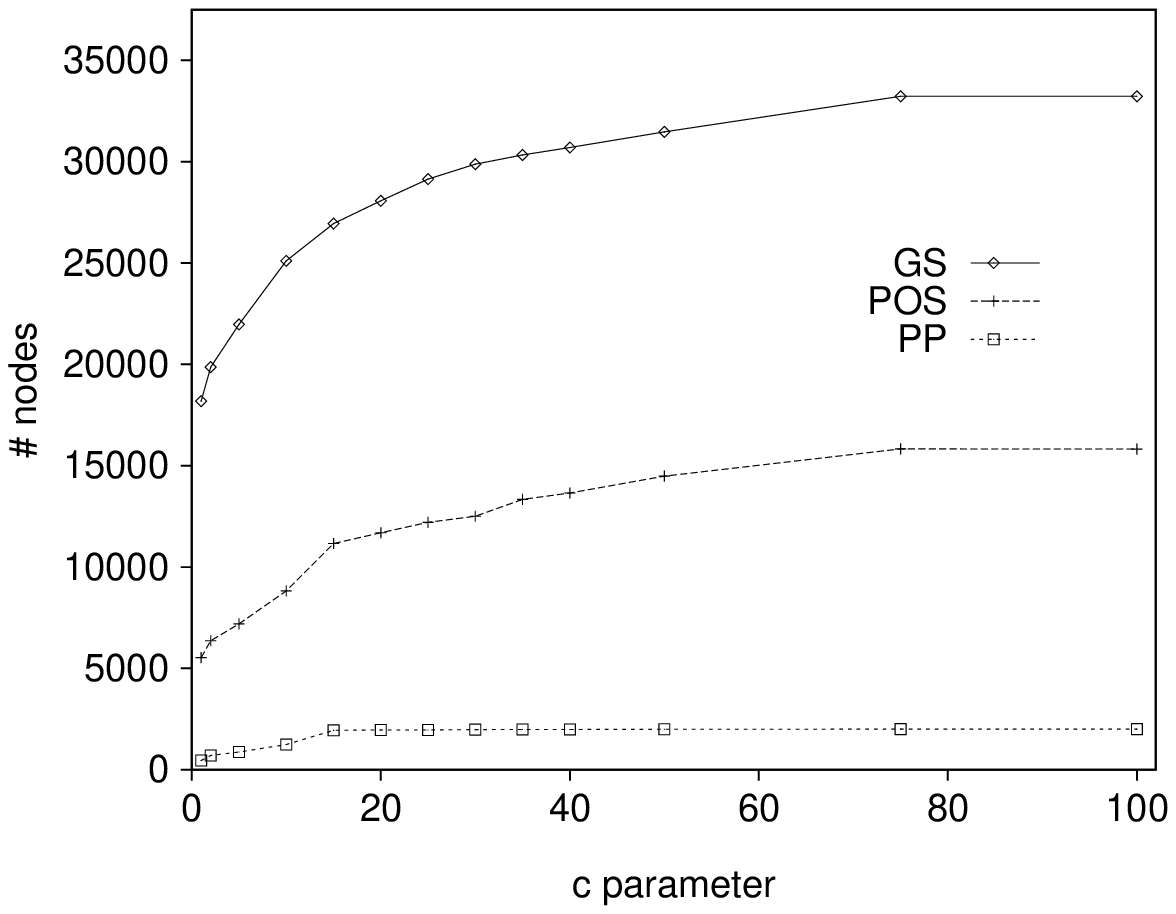}
        \epsfxsize=0.5\textwidth
        \epsfbox{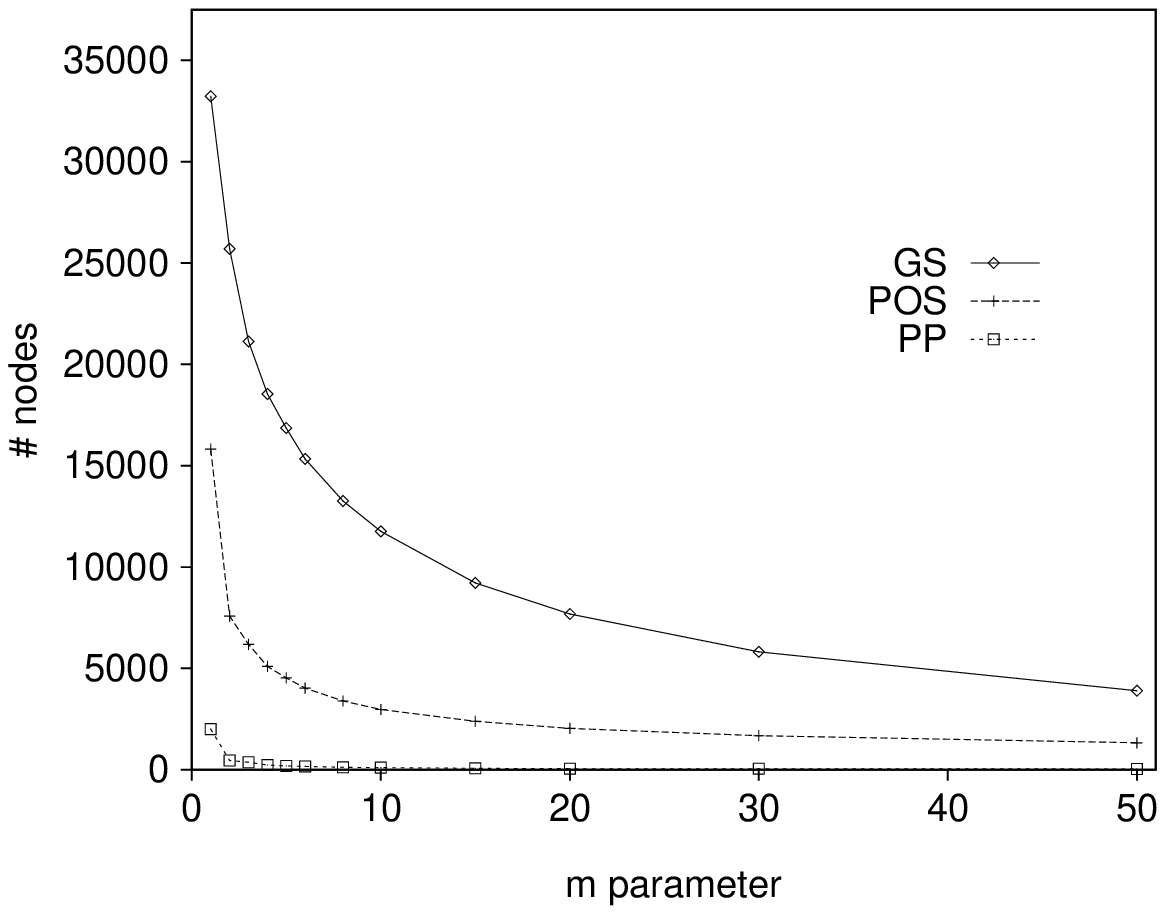}
}
\caption{Tree sizes (number of nodes) generated by {\sc c5.0} with
increasing $c$ parameter (left) and increasing $m$ parameter (right),
for the {\sc gs}, {\sc pos}, and {\sc pp} tasks.}
\label{c50-trees-perf}
\end{figure}

Figure~\ref{c50-parameter-perf} displays the effect on generalization
accuracy of varying the $c$ parameter from 1 to 100 (left) and the $m$
parameter from 1 to 50 (right).  Performance of {\sc c5.0} on the {\sc
pos} and {\sc pp} tasks is only slightly sensitive to the setting of
both parameters, while the performance on the {\sc gs} task is
seriously harmed when $c$ is too small (i.e., when pruning is high),
or when $m$ is larger than 1 (i.e., when single instances to be
disambiguated are ignored). The direct effect of changing both
parameters is shown in Figure~\ref{c50-trees-perf}; small values of
$c$ lead to smaller trees, as do large values of $m$. For the {\sc
pos}, and {\sc pp} tasks, it is interesting to note that the
performance of {\sc c5.0}, although usually lower than that of {\sc
ib1-ig}, is maintained even with a small number of nodes: with $m=50$
and $c=100$, {\sc c5.0} needs 1324 nodes for the {\sc pos} task and 34
nodes for the {\sc pp} task. However, nodes in these trees contain a
lot of information since grouping of feature values was used.

Table \ref{optvsdef} compares {\sc c5.0} with default settings ({\sc
c5.0def}) to {\sc c5.0} with `lazy' parameter setting $c=100$ and
$m=1$ ({\sc c5.0lazy}). The differences are significant at the
$p<0.05$ level for the {\sc gs} and {\sc pos} data sets, but not for
the {\sc pp} data set.

\begin{table}
\begin{center}
\begin{tabular}{|l|rr|rr|}
\hline
 & \multicolumn{4}{|c|}{Generalization accuracy} \\
 & \multicolumn{2}{|c|}{\sc c5.0lazy} & 
   \multicolumn{2}{|c|}{\sc c5.0def} \\
Task & \% & $\pm$ & \% & $\pm$ \\
\hline
{\sc gs}  & 93.34 & 0.13 & 92.48 & 0.14 \\
{\sc pos} & 97.92 & 0.04 & 97.97 & 0.04 \\
{\sc pp}  & 80.85 & 1.07 & 80.89 & 1.01 \\
\hline
\end{tabular}
\caption{10 fold CV generalization accuracies (in terms of percentages
of correctly classified test instances) on the {\sc gs}, {\sc pos},
and {\sc pp} tasks, by {\sc c5.0} with parameter setting $c=25$ and
$m=2$ (default setting), and {\sc c5.0} with parameter setting $c=100$
and $m=1$ (`lazy' setting).}
\label{optvsdef}
\end{center}
\end{table}

These parameter tuning results indicate that decision-tree pruning is
not beneficial to generalization accuracy, but neither is it generally
harmful. Only on the {\sc gs} task are strong decreases in generalization
accuracy found with decreasing $c$. Likewise, small decreases in
performance are witnessed with increasing $m$ for the {\sc pos} and
{\sc pp} tasks, while a strong accuracy decrease is found with
increasing $m$ for the {\sc gs} task.

\subsubsection{Efficiency}

In addition to generalization accuracy, which is the focus of our
attention in this research, efficiency, measured in terms of training
and testing speed and in terms of memory requirements, is also an
important criterion to evaluate learning algorithms. For training,
{\sc ib1-ig} is fastest as it reduces to storing instances and
computing information gain (although in the implementation we used,
various indexing strategies are used), and {\sc c5.0}, because of the
computation involved in recursively partitioning the training set,
value grouping, and pruning, is the slowest. {\sc igtree} occupies a
place in between, similar to {\sc ib1-ig} in training time. Memory
requirements are, in theory, highest in {\sc ib1-ig} and lowest for
{\sc c5.0} with default parameter settings. Again, {\sc igtree} is in
between, similar to {\sc c5.0} in memory usage. However, in practice,
the implementations of {\sc c5.0} and {\sc igtree} store the entire
data set during training and hence take up more space than {\sc
ib1-ig}. Finally, for testing speed, the most important efficiency
measurement, {\sc igtree} and {\sc c5.0} are on a par, and both are
some 2 orders of magnitude faster than {\sc ib1-ig}. In
\namecite{Daelemans+97}, the asymptotic complexity of {\sc ib1-ig} and
{\sc igtree} is described. Illustrative timing results on the first
partition of each of the data sets are provided in Table~\ref{timing}.
See~\namecite{Daelemans+98} for the details of the effects of various
optimizations in the TiMBL package.

\begin{table}
\begin{center}
\begin{tabular}{|l|rrr|rrr|rrr|}
\hline
 & \multicolumn{9}{|c|}{Time (seconds)} \\
\hline
Task  & \multicolumn{3}{|c|}{\sc c5.0} & \multicolumn{3}{|c|}{\sc igtree} & \multicolumn{3}{|c|}{\sc ib1-ig}\\
      & train & test & total & train & test & total & train & test & total\\
\hline
{\sc gs} & - & - & 2406    & 79  & 9  & 88 & 83 & 2391 & 2474\\
{\sc pos} & - & - & 7234   & 43  & 18 & 61 & 211& 6416 & 6627\\
{\sc pp} & - & - & 295     & 6   & 1  &  7 &  7 & 10   & 17\\
{\sc np} & - & - & - & 152 & 8  &160 & 98 & 19474& 19572\\
\hline
\end{tabular}
\caption{Timing results in seconds (elapsed wall clock time) for the
first partition of all four data sets, measured on a SUN Sparc 5 with
160 MB internal memory. The results for {\sc c5.0} were obtained
through its own internal timer which does not differentiate between
training and testing time. The results for {\sc ib1-ig} and {\sc
igtree} were obtained using TiMBL and its internal timer.}
\label{timing}
\end{center}
\end{table}

In this Section, we have shown that when comparing the generalization
accuracy of {\sc ib1-ig} to that of decision tree methods, we see the
same results as in our experiments on editing: different types of
abstraction (some of them explicitly aimed at removing exceptional
instances) do not succeed in general in providing a better
generalization accuracy than {\sc ib1-ig}.  However, for some
data sets, if a lower generalization accuracy is acceptable, the
pruning and abstraction methods of {\sc c5.0} are able to induce
compact decision trees without a significant loss in initial
generalization accuracy.

\section{Why forgetting exceptions is harmful}
\label{why}

In this section we explain why forgetting exceptional instances,
either by editing them from memory or by pruning them from decision
trees, is harmful to generalization accuracy for the language
processing tasks studied.  We explain this effect on the basis of the
properties of this type of task and the properties of the learning
algorithms used. Our approach of studying data set properties, to find
an explanation for why one type of inductive algorithm rather than
another is better suited for learning a type of task, is in the spirit
of \namecite{Aha92} and \namecite{Michie+94}.

\subsection{Properties of language processing tasks}

Language processing tasks are usually described as complex mappings
between representations: from spelling to sound, from strings of words
to parse trees, from parse trees to semantic formulas, etc.  These
mappings can be approximated by (cascades of) classification tasks
\cite{Ratnaparkhi97b,Daelemans96,Cardie96,Magerman94} which makes them
amenable to machine learning approaches.  One of the most salient
characteristics of natural language processing mappings is that they
are noisy and complex. Apart from some regularities, they contain also
many sub-regularities and (pockets of) exceptions. In other words,
apart from a core of generalizable regularities, there is a relatively
large periphery of irregularities \cite{Daelemans96}.  In rule-based
{\sc nlp}, this problem has to be solved using mechanisms such as rule
ordering, subsumption, inheritance, or default reasoning (in
linguistics this type of ``priority to the most specific'' mechanism
is called the {\em elsewhere condition}).  In the feature-vector-based
classification approximations of these complex language processing
mappings, this property is reflected in the high degree of
disjunctivity of the instance space: classes exhibit a high degree of
polymorphism.  Another issue we study in this Section is the
usefulness of exceptional as opposed to more regular instances in
classification.

\subsubsection{Degree of polymorphism}

\begin{figure}
\centerline{
        \epsfxsize=0.8\textwidth
        \epsfbox{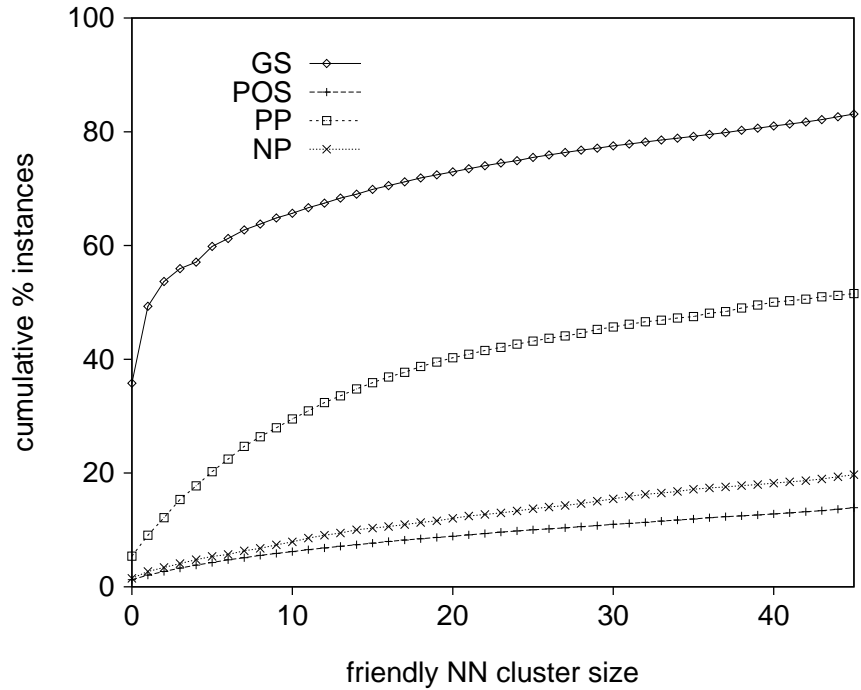} }
\caption{Cumulative percentages of occurrences of
friendly-neighbor clusters of sizes 0 to 45, as found in
the {\sc gs}, {\sc pos}, {\sc pp}, and {\sc np} data sets.}
\label{iblclusters}
\end{figure}

Several quantitative measures can be used to show the degree of
polymorphism: the number of clusters (i.e., groups of nearest-neighbor
instances belonging to the same class), the number of disjunct
clusters per class (i.e., the numbers of separate clusters per class),
or the numbers of prototypes per class \cite{Aha92}.  We approach the
issue by looking at the average number of friendly neighbors per
instance in a leave-one-out experiment \cite{Weiss+91}.  For each
instance in the four data sets a distance ranking of the 50 nearest
neighbors to an instance was produced. In case of ties in distance,
nearest neighbors with an identical class as the left-out instance are
placed higher in rank than instances with a different class. Within
this ranked list we count the ranking of the nearest neighbor of a
different class. This rank number minus one is then taken as the
cluster size surrounding the left-out instance. If, for example, a
left-out instance is surrounded by three instances of the same class
at distance 0.0 (i.e., no mismatching feature values), followed by a
fourth nearest-neighbor instance of a different class at distance 0.3,
the left-out instance is said to be in a cluster of size three. The
results of the four leave-one-out experiments are displayed
graphically in Figure~\ref{iblclusters}.  The $x$-axis of
Figure~\ref{iblclusters} denotes the numbers of friendly neighbors
found surrounding instances; the $y$-axis denotes the cumulative
percentage of occurrences of friendly-neighbor clusters of particular
sizes.

The cumulative percentage graphs in Figure~\ref{iblclusters} display 
that for the case of the {\sc gs} task, many instances have only a 
handful of friendly neighbors; 59.9\% of the {\sc gs} instances have 
five friendly neighbors or less, while 35.8\% has no friendly 
neighbors at all.  For the case of the {\sc pp} task, the number of 
friendly neighbors is larger; 50.1\% of the {\sc pp} instances have 40 
or less friendly neighbors.  Instances of the {\sc pos} and {\sc np} 
tasks tend to have even more friendly neighbors surrounding them.  In 
sum, the {\sc gs} task appears to display high disjunctivity (i.e., a 
high degree of polymorphism) of its 159 classes; for the other three 
tasks, disjunctivity appears to be slightly lower, but still the 
classes are scattered across many unconnected clusters in the instance 
space.

In sum, we find indications for a high disjunctity or polymorphism of
the language data sets investigated in this study. Other studies in
which machine learning algorithms are applied to language data, and in
which special attention is payed to learning exceptions, mention
similar indications (e.g.,
\namecite{Mooney+95,VandenBosch+95}). However, the question whether
language data in general exhibits a higher degree of disjunctiveness
or polymorphism than comparable data sets of non-linguistic origin
remains an open one, and will be a focal point in future research.

\subsubsection{Usefulness of exceptional instances}

\begin{figure}
        \begin{center}
	\begin{minipage}[c]{0.45\textwidth}
		\epsfxsize=\textwidth
		\epsffile{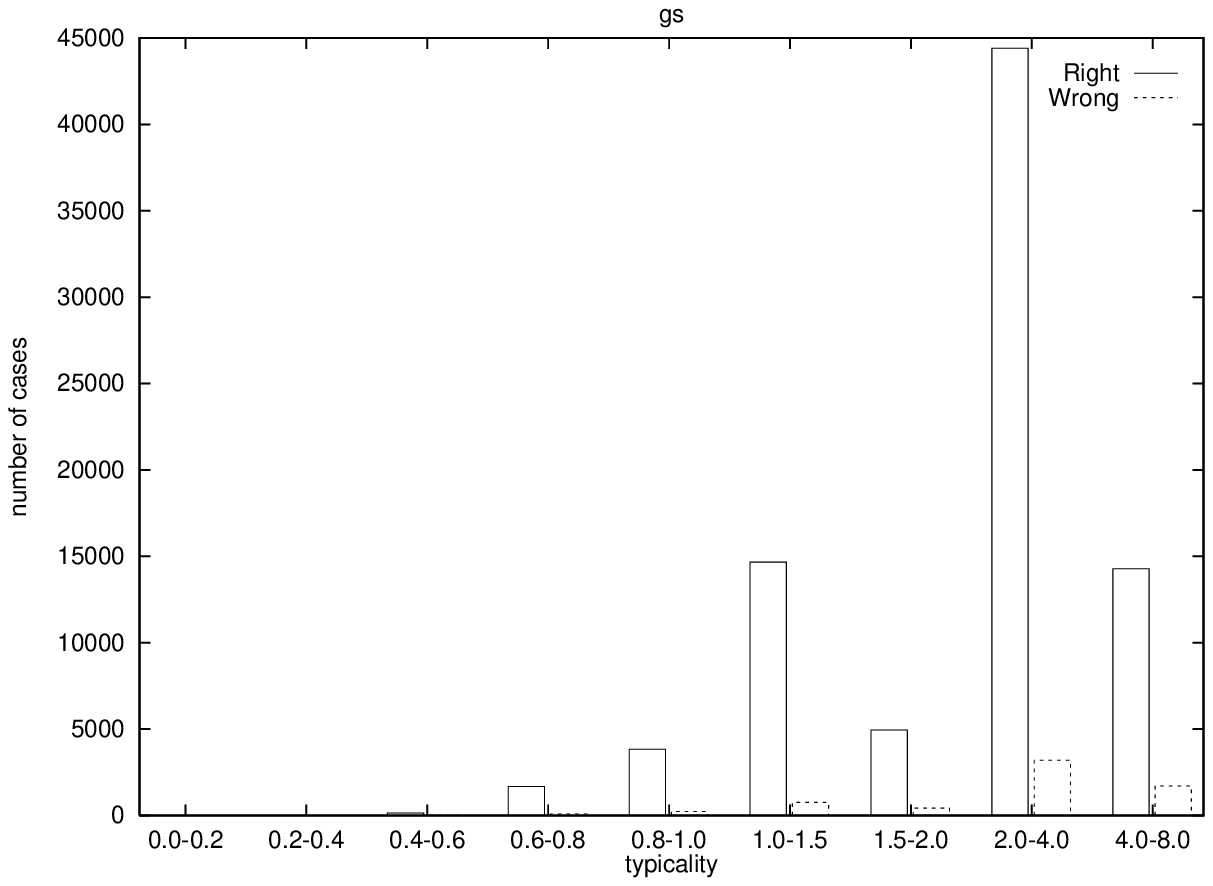}
	\end{minipage}%
	\begin{minipage}[c]{0.45\textwidth}
		\epsfxsize=\textwidth
		\epsffile{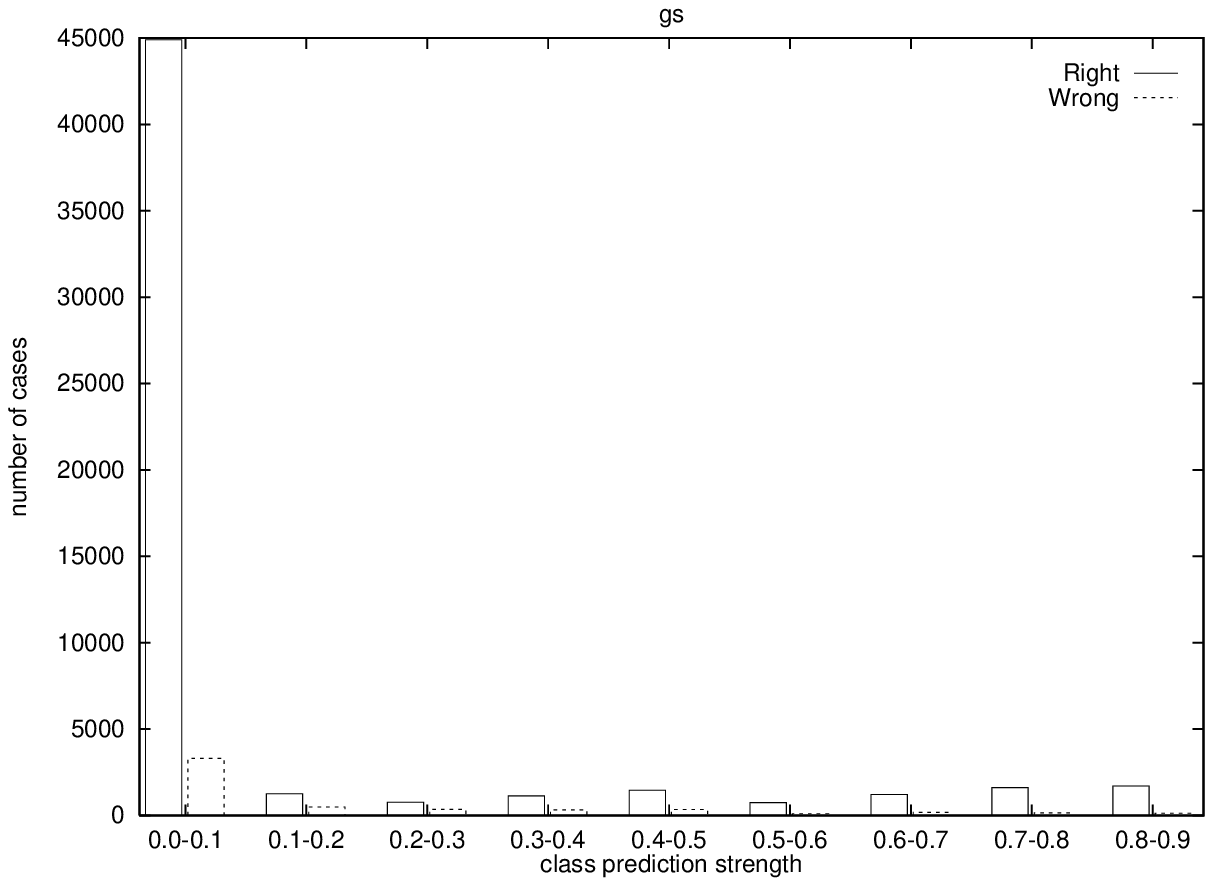}
	\end{minipage}
	\begin{minipage}[c]{0.45\textwidth}
		\epsfxsize=\textwidth
		\epsffile{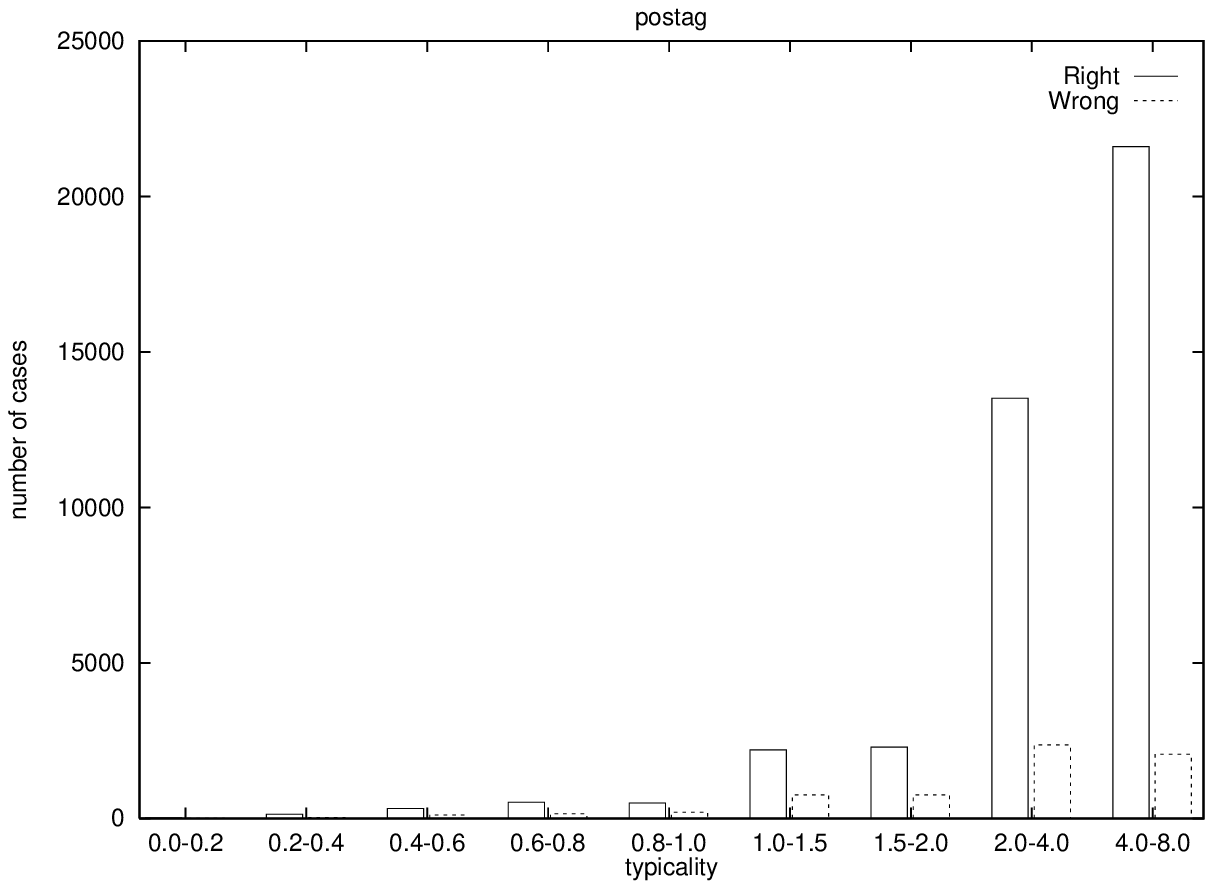}
	\end{minipage}%
	\begin{minipage}[c]{0.45\textwidth}
		\epsfxsize=\textwidth
		\epsffile{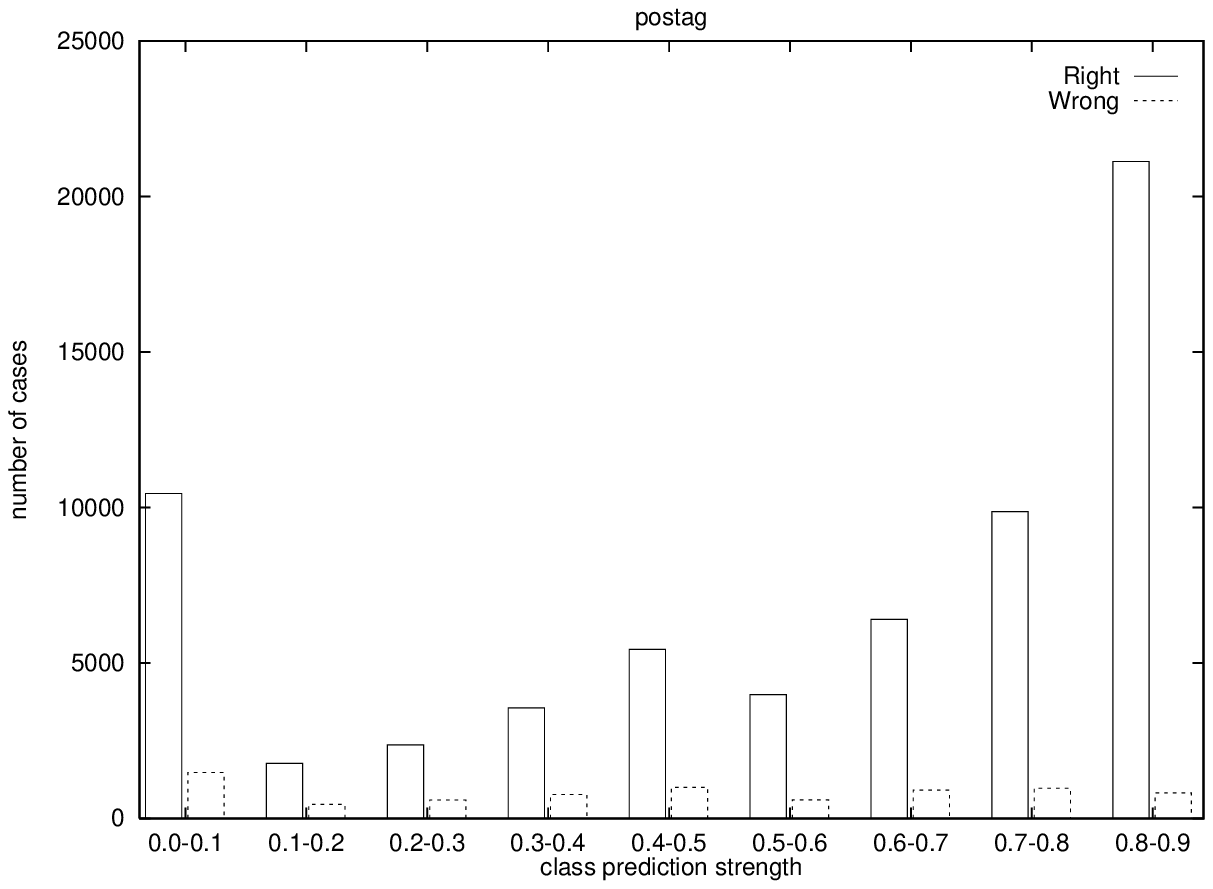}
	\end{minipage}
	\begin{minipage}[c]{0.45\textwidth}
		\epsfxsize=\textwidth
		\epsffile{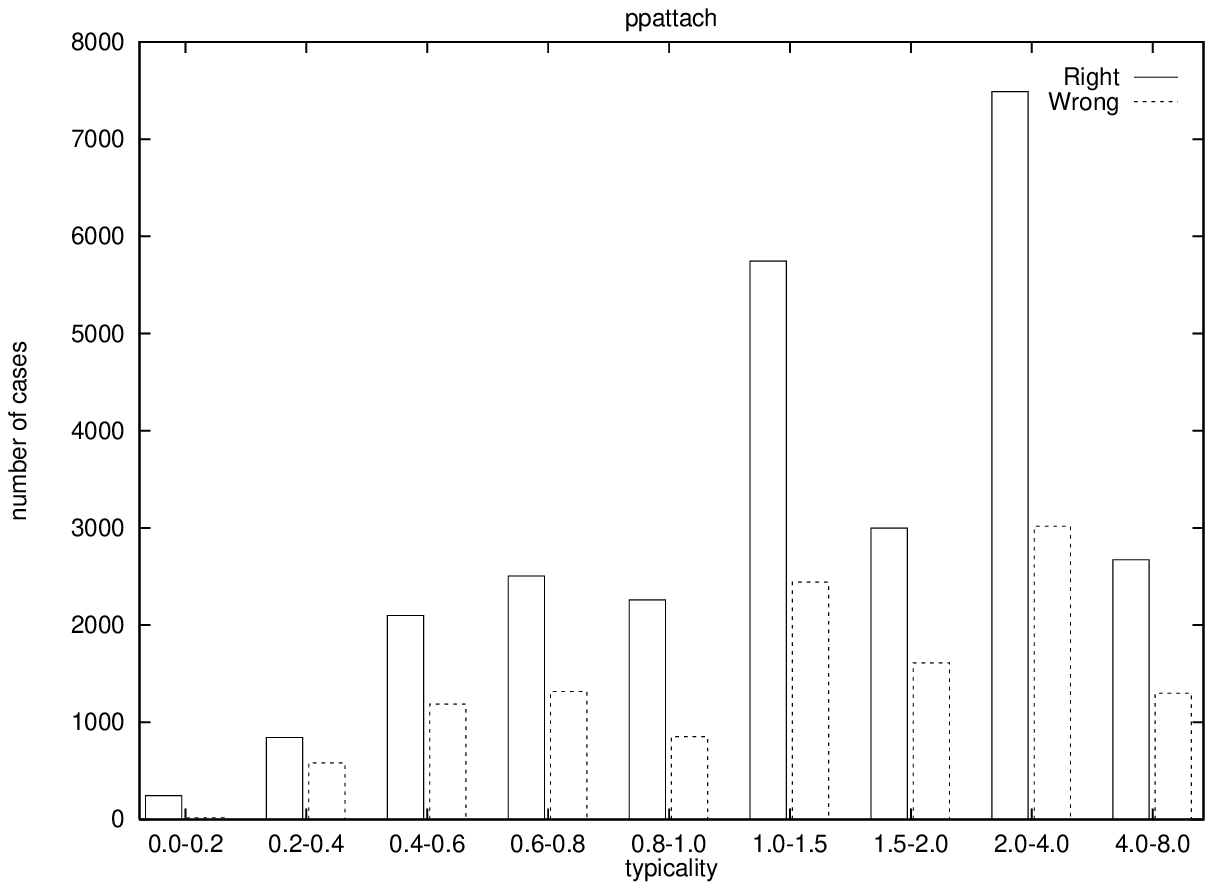}
	\end{minipage}%
	\begin{minipage}[c]{0.45\textwidth}
		\epsfxsize=\textwidth
		\epsffile{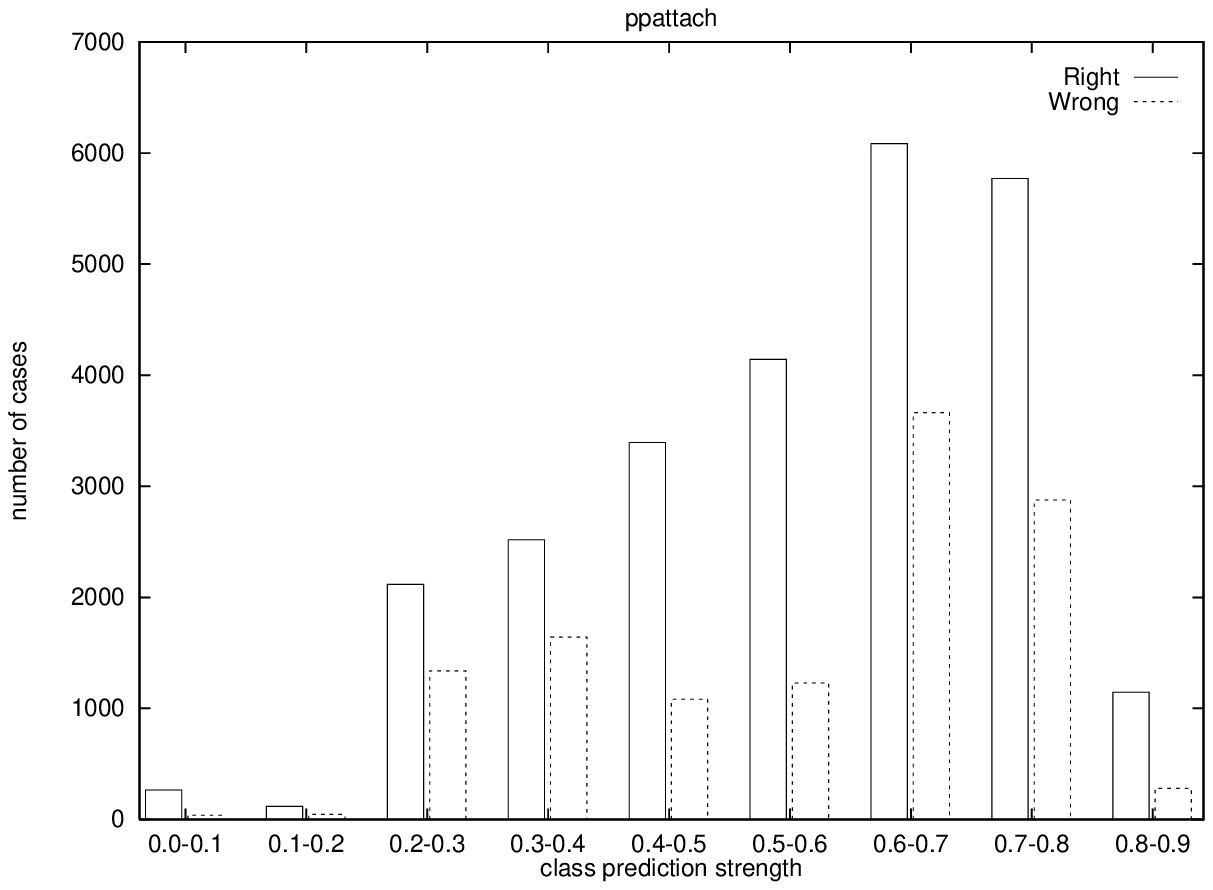}
	\end{minipage}
	\begin{minipage}[c]{0.45\textwidth}
		\epsfxsize=\textwidth
		\epsffile{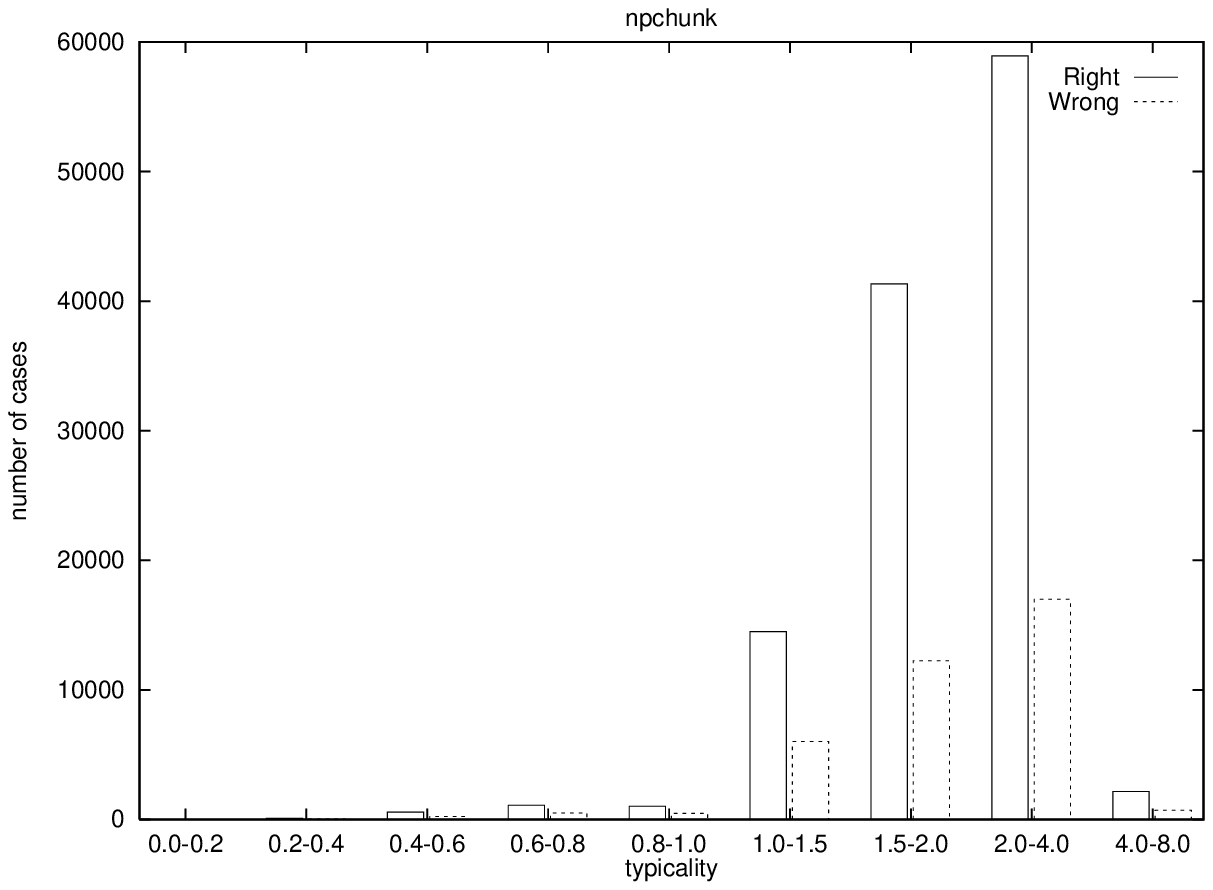}
	\end{minipage}%
	\begin{minipage}[c]{0.45\textwidth}
		\epsfxsize=\textwidth
		\epsffile{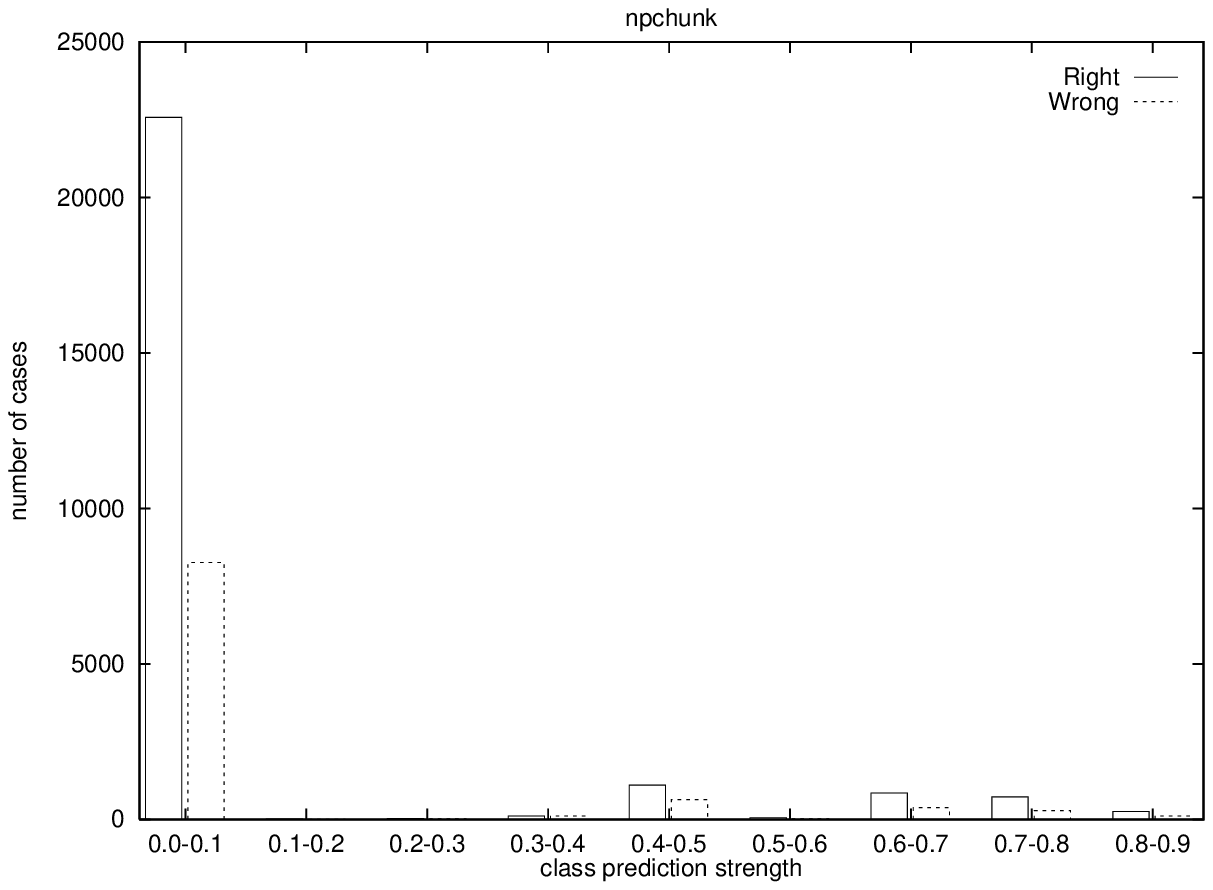}
	\end{minipage}
		\caption{Histograms per typicality (left) and  class-prediction strength (right)
                of the neighbors present in support sets for each of
		the four tasks. For each range (indicated at the
		$x$-axes), the number of instances leading to a
		correct classification (Right), and to a
		misclassification (Wrong), is displayed as a bar.}
		\label{cps-radio}
	\end{center}
\end{figure}

Having established a fairly high degree of disjunctivity for our data
sets, an indication is needed that fully retaining this disjunctivity
is indeed beneficial. With this in mind, we can return to our editing
experiments and examine why even instances with low typicality or low
prediction strength cannot be removed from the training data.  For
this purpose, we have looked at the instances that are actually used
in the memory-based classification process to classify the test
instances. We call the nearest neighbors that were used to classify
test instances the {\em support set}. The distribution of both
typicality and {\sc cps} over the support set can be
seen in Figure~\ref{cps-radio}. The support set can be divided into
support for correct decisions ({\em Right}) and errors ({\em
Wrong}). The average number of neighbors for correct decisions is
approximately the same as for errors. The figures clearly show that
even instances with respectively low typicality (below 1.0) or low
{\sc cps} (below 0.5) are more often used to support
correct decisions than errors. Although this does not present a proof
of the detrimental effects of their removal, it does show that
exceptional events can be beneficial for accurate generalization. The
small disjunctive clusters are productive for classifying new
instances.

\subsection{Properties of learning algorithms}

If we classify instance $X$ by looking at its nearest neighbors, we 
are in fact estimating the probability $P(class|X)$, by looking at the 
relative frequency of the class in the set defined by $sim_{k}(X)$, 
where $sim_{k}(X)$ is a function from $X$ to the set of most similar 
instances present in the training data.  The $sim_{k}(X)$ function 
given by the overlap metric groups varying numbers of instances into 
{\em buckets}\/ of equal similarity.  A bucket is defined by a 
particular number of mismatches with respect to instance $X$.  Each 
bucket can further be decomposed into a number of {\em schemata}\/ 
characterized by the position of the mismatch.

The search for the nearest neighbors results in the use of the most
similar instantiated schema or bucket for extrapolation.  In
statistical language modeling this is known as backed-off
estimation~\cite{Collins+95,Katz87}.  The distance metric defines a
specific-to-general ordering ($X \prec Y$: read $X$ is more specific
than $Y$, see also~\namecite{Zavrel+97}), where the most specific
schema is the schema with zero mismatches (i.e., an identical instance
in memory), and the most general schema has a mismatch on every
feature, which corresponds to the entire memory being retrieved.

If information gain weights are used in combination with the overlap
metric, individual schemata instead of buckets become the steps of the
back-off sequence (unless two schemata are exactly tied in their IG
values).  The $\prec$ ordering becomes slightly more complicated now,
as it depends on the number of wild-cards {\em and}\/ on the magnitude
of the weights attached to those wild-cards.  Let $S$ be the most
specific (zero mismatches) schema. We can then define the $\prec$
ordering between schemata in the following equation, where
$\Delta(X,Y)$ is the distance as defined in Equation~\ref{distance}.
\begin{equation}
S' \prec S'' \ \Leftrightarrow \ \Delta(S,S') < \Delta(S,S'')
\label{subset_with_IG}
\end{equation}
This approach represents a type of implicit parallelism. The
importance of all of the $2^{F}$ schemata is specified using only $F$
parameters (i.e., the IG weights), where $F$ is the number of
features.  Moreover, using the schemata keeps the information from all
training instances available for extrapolation in those cases where
more specific information is not available.

Decision trees can also be described as backed-off estimators of the
class probability conditioned on the combination of the
features-values. However, here some schemata are not available for
extrapolation. Even in a decision tree without any pruning, such
abstraction takes place. Once a test instance matches an arc with a
certain value for a particular feature, the set of schemata from which
it can receive a classification is restricted to those for which that
feature matches. This means that other schemata which are more
specific when judged by the ordering of Equation~\ref{subset_with_IG},
are unavailable. If pruning is applied, even more schemata are
blocked.

\begin{figure}
        \begin{center}
		\leavevmode
                \epsfxsize=11cm
                \epsfysize=7cm
                \epsffile{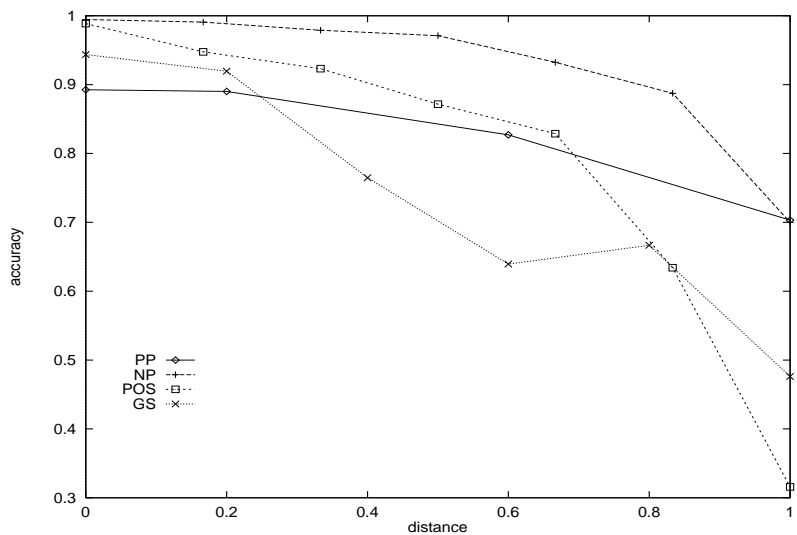}
		\vspace*{-4mm}
		\caption{Percentage correct for our data sets 
		plotted as a function of distance between the test
		instance and its nearest neighbor. The distances are
		normalized between
		zero and one, and discretized into a maximum of ten
		evenly spaced intervals to make a comparison across data sets possible.
		}
		\label{spec-vs-corr}
	\end{center}
\end{figure}

Figure~\ref{spec-vs-corr} shows why this elimination of schemata can
be harmful. In this figure the percentage correct for our data sets is
plotted as a function of specificity. The decrease of the accuracy
seen in the graph clearly confirms the intuition that an extrapolation
from a more specific support set is more likely to be
correct. Reasoning in the other direction, it suggests that any
forgetting of specific information from the training set will push at
least some test instances in the direction of a less specific support set,
and thus of lower accuracy. 

\begin{table}
\begin{center}
\setlength{\tabcolsep}{0.8mm}
\begin{tabular}{|l|rrc|rrc|rrc|rrc|}
\hline
 & \multicolumn{12}{|c|}{Average IG Overlap Distance (number of instances)} \\
\hline
Task  & \multicolumn{3}{|c|}{FF} & \multicolumn{3}{|c|}{FT} & \multicolumn{3}{|c|}{TF} & \multicolumn{3}{|c|}{TT}\\
      & {\sc ib1} & {\sc igt} & n & {\sc ib1} & {\sc igt} & n & {\sc ib1} & {\sc igt} & n & {\sc ib1} & {\sc igt}& n\\
\hline
{\sc gs} & 0.03 & 0.05 & (4083) & 0.08 & 0.14 & (249) & 0.10 & 0.19 & (552) & 0.01 & 0.02 & (62633)\\
{\sc pos} & 0.18 & 0.23 & (1876) & 0.26 & 0.37 & (440) & 0.27 & 0.40 & (524) & 0.07 & 0.08 & (101776)\\
{\sc pp} & 0.06 & 0.07 & (275) & 0.06 & 0.08 & (111) & 0.06 & 0.07 & (184) & 0.05 & 0.06 & (1820)\\
{\sc np} & 0.12 & 0.19 & (343) & 0.14 & 0.24 & (160) & 0.14 & 0.26 & (324) & 0.08 & 0.15 & (24286)\\
\hline
\end{tabular}
\caption{The average distance at which classification takes place for
{\sc ib1-ig} (listed under {\sc ib1}) and {\sc igtree} (listed under
{\sc igt}). The distances have been split out into four conditions:
FF, FT, TF, and TT; the first letter refers to {\sc ib1-ig} giving a
False or True answer, the second refers in the same manner to the
output of {\sc igtree}. The third column gives the number of instances for
that condition. The {\sc igtree} distances have been computed from an
{\em unpruned}\/ tree.}
\label{reachability}
\end{center}
\end{table}

A more direct illustration of this matter can be given for the limited
accessibility of schemata in {\sc igtree}. As the ordering of features
is constant throughout the tree, the schemas that are accessible at any
given node in the tree are limited to those that match all features
with a higher {\sc ig} weight. The depth of the {\sc igtree} node at
which classification was performed can directly be translated into a
distance between the test pattern and the branch of the tree, using
the {\sc ig} weights. To make the comparison fair, we have used an
{\em unpruned}\/ {\sc igtree}. Table~\ref{reachability} shows the
average distances at which classifications were made for the four
tasks at hand. {\sc igtree} consistently classifies at a larger
average distance than {\sc ib1-ig}. Moreover, through analysis of
those test instances that were misclassified by {\sc igtree}, but
classified correctly by {\sc ib1-ig} (i.e., TF in
Table~\ref{reachability}), we found that for a majority (69\% for {\sc
gs}, 90\% for {\sc pos}, 55\% for {\sc pp}, and 100\% for {\sc np}) of
these instances the classification distance was larger for {\sc
igtree} than for {\sc ib1-ig}. This means that in all these cases a
closer neighbor was available to support a correct classification, but
was not used, because its schema was not accesible.

\subsubsection{Increasing $k$}

As an aside, we note that we have reported solely on experiments with
{\sc ib1-ig} with $k=1$.  Although it is not directly related to
``forgetting'', taking a larger value of $k$ can also be considered as
a type of abstraction, because the class is estimated from a somewhat
smoothed region of the instance space.  Only on the basis of the
results described so far, we cannot claim that $k=1$ is the optimal
setting for our experiments.  The results discussed above suggest that
the average `$k$' actually surrounding an instance is larger than $1$,
although many instances have only one or no friendly neighbor,
especially in the case of the {\sc gs} task.  The latter suggests that
a considerable amount of ambiguity is found in instances that are
highly similar; matching with $k>1$ may fail to detect those cases in
which an instance has one best-matching friendly neighbor, and many
next-best-matching instances of a different class.

\begin{table}
\begin{center}
\begin{tabular}{|l|rrrr|}
\hline
 & \multicolumn{4}{|c|}{Generalization accuracy (\%)} \\
Task  & $k=1$ & $k=2$ & $k=3$ & $k=5$ \\
\hline
{\sc gs}  & 93.45 $\pm$ 0.15 & 93.00 $\pm$ 0.15 & 92.71 $\pm$ 0.13 & 92.30 $\pm$ 0.12\\
{\sc pos} & 97.86 $\pm$ 0.05 & 97.72 $\pm$ 0.05 & 97.27 $\pm$ 0.04 & 95.91 $\pm$ 0.05 \\
{\sc pp}  & 83.48 $\pm$ 1.16 & 78.10 $\pm$ 1.26 & 75.19 $\pm$ 1.75 & 75.67 $\pm$ 1.53 \\
{\sc np}  & 98.07 $\pm$ 0.05 & 98.05 $\pm$ 0.05 & 98.23 $\pm$ 0.07 & 98.15 $\pm$ 0.09 \\
\hline
\end{tabular}
\caption{Generalization accuracies (in terms of percentages of
correctly classified test instances) on the {\sc gs}, {\sc pos}, {\sc
pp}, and {\sc np} tasks, by {\sc ib1-ig} with $k=1$, 2, 3, and 5.}
\label{ib1ig-k}
\end{center}
\end{table}

We performed experiments with {\sc ib1-ig} on the four tasks with
$k=2$, $k=3$, and $k=5$, and mostly found a decrease in generalization
accuracy. Table~\ref{ib1ig-k} lists the effects of the higher values
of $k$. For all tasks except {\sc np}, setting $k>1$ leads to a
harmful abstraction from the best-matching instance(s) to a more
smoothed best matching group of instances.

In this Section, we have tried to interpret our empirical results in 
terms of properties of the data and of the learning algorithms used.  
A salient characteristic of our language learning tasks, shown most 
clearly in the {\sc gs} data set but also present in the other 
data sets, is the presence of a high degree of class polymorphism (high 
disjunctivity).  In many cases, these small disjuncts constitute 
productive (pockets of) exceptions which are useful in producing 
accurate extrapolations to new data.  {\sc ib1-ig}, through its 
implicit parallelism and its feature relevance weighting, is better 
suited than decision tree methods to make available the most specific 
relevant patterns in memory to extrapolate from.

\section{Related research}

\namecite{Daelemans95} provides an overview of memory-based learning
work on phonological and morphological tasks (grapheme-to-phoneme
conversion, syllabification, hyphenation, morphological synthesis,
word stress assignment) at Tilburg University and the University of
Antwerp in the early nineties.  The present paper directly builds on
the results obtained in that research.  More recently, the approach
has been applied to part-of-speech tagging (morphosyntactic
disambiguation), morphological analysis, and the resolution of
structural ambiguity (prepositional-phrase attachment)
\cite{Daelemans+96,VandenBosch+96,Zavrel+97b}.  Whenever these studies
involve a comparison of memory-based learning to more eager methods, a
clear advantage of memory-based learning is reported.

Cardie \shortcite{Cardie93,Cardie94} suggests a memory-based learning 
approach for both (morpho)syntactic and semantic disambiguation and 
shows excellent results compared to alternative approaches.  
\namecite{Ng+96} report results superior to previous statistical 
methods when applying a memory-based learning method to word sense 
disambiguation.  In reaction to \namecite{Mooney96} where it was shown 
that naive Bayes performed better than memory-based learning, 
\namecite{Ng97b} showed that with higher values of $k$, memory-based 
learning obtained the same results as naive Bayes.

The exemplar-based reasoning aspects of memory-based learning are also 
prominent in the large literature on example-based machine translation 
(cf.  \namecite{Jones96} for an overview), although systematic 
comparisons to eager approaches seem to be lacking in that field.

In the recent literature on statistical language learning, which 
currently still largely adheres to the hypothesis that what is 
exceptional (improbable) is unimportant, similar results as those 
discussed here for machine learning have been reported.  In 
\namecite{Bod95}, a data-oriented approach to parsing is described in 
which a treebank is used as a `memory' and in which the parse of a new 
sentence is computed by reconstruction from subtrees present in the 
treebank.  It is shown that removing all hapaxes (unique subtrees) 
from memory degrades generalization performance from 96\% to 92\%.  
Bod notes that ``this seems to contradict the fact that probabilities 
based on sparse data are not reliable.''  (\namecite{Bod95}, p.68).  
In the same vein, \namecite{Collins+95} show that when applying the 
back-off estimation technique \cite{Katz87} to learning 
prepositional-phrase attachment, removing all events with a frequency 
of less than 5 degrades generalization performance from 84.1\% to 
81.6\%.  In \namecite{Dagan+97}, finally, a similarity-based 
estimation method is compared to back-off and maximum-likelihood 
estimation on a pseudo-word sense disambiguation task.  Again, a 
positive effect of events with frequency 1 in the training set on 
generalization accuracy is noted.  

In the context of statistical language learning, it is also relevant
to note that as far as comparable results are available,
statistical techniques, which also abstract from exceptional events,
never obtain a higher generalization accuracy than {\sc ib1-ig}
\cite{Daelemans95,Zavrel+97,Zavrel+97b}. Reliable comparisons (in the
sense of methods being compared on the same train and test data) with
the empirical results reported here cannot be made, however.

In the machine learning literature, the problem of {\em small
disjuncts}\/ in concept learning has been studied before by
\namecite{Quinlan91}, who proposed more accurate error estimation
methods for small disjuncts, and by \namecite{Holte+89}.  The latter
define a small disjunct as one that has small coverage (i.e., a small
number of training items are correctly classified by it).  This
definition differs from ours, in which small disjuncts are those that
have few neighbors with the same category.  Nevertheless, similar
phenomena are noted: sometimes small disjuncts constitute a
significant portion of an induced definition, and it is hard to
distinguish productive small disjuncts from noise (see also
\namecite{Danyluk+93}).  A maximum-specificity bias for small
disjuncts is proposed to make small disjuncts less error-prone.
Memory-based learning is of course a good way of implementing this
remedy (as noted, e.g., in \namecite{Aha92}).  This prompted
\namecite{Ting94} to propose a composite learner with an
instance-based component for small disjuncts, and a decision tree
component for large disjuncts.  This hybrid learner improves upon the
{\sc c4.5} baseline for several definitions of `small disjunct' for
most of the data sets studied.  Similar results have recently been
reported by \namecite{Domingos96}, where {\sc rise}, a unification of
rule induction ({\sc c4.5}) and instance-based learning ({\sc pebls})
is proposed.  In an empirical study, {\sc rise} turned out to be
better than alternative approaches, including its two `parent'
algorithms.  The fact that rule induction in {\sc rise} is
specific-to-general (starting by collapsing instances) rather than
general-to-specific (as in the decision tree methods used in this
paper), may make it a useful approach for our language data as well.

\section{Conclusion and future research}

We have provided empirical evidence for the hypothesis that forgetting 
exceptional instances, either by editing them away according to some 
exceptionality criterion in memory-based learning or by abstracting 
from them in decision-tree learning, is harmful to generalization 
accuracy in language learning.  Although we found some exceptions to 
this hypothesis, the fact that abstraction or editing is {\em never 
beneficial}\/ to generalization accuracy is consistently shown in all 
our experiments.

Data sets representing {\sc nlp} tasks show a high degree of 
polymorphism: categories are represented in instance space as small 
regions with the same category separated by instances with a different 
category (the categories are highly disjunctive).  This was 
empirically shown by looking at the average number of friendly 
neighbors per instance; an indirect measure of the average size of the 
homogeneous regions in instance space.  This analysis showed that for 
our {\sc nlp} tasks, classes are scattered across many disjunctive 
clusters in instance space.  This turned out to be the case especially 
for the {\sc gs} data set, the only task presented here which has 
extensively been studied in the {\sc ML} literature before (through 
the similar {\sc nettalk} data set).  It will be necessary to 
investigate polymorphism further using more language data sets and more 
ways of operationalizing the concept of `small disjuncts'.

The high disjunctivity explains why editing the training set in
memory-based learning using typicality and {\sc cps} criteria does not
improve generalization accuracy, and even tends to decrease it.  The
instances used for {\em correct}\/ classification (what we called the
support set) are as likely to be low-typical or
low-class-prediction-strength (thus exceptional) instances as
high-typical or high-class-prediction-strength instances.  The editing
that we find to be the most harmless (although never beneficial) to
generalization accuracy is editing up to about 20\% high-typical and
high-class-prediction-strength instances.  Nevertheless, these results
leave room for combining memory-based learning and specific-to-general
rule learning of the kind presented in \namecite{Domingos96}.  It
would be interesting further research to test his approach on our
data.

The fact that the generalization accuracies of the decision-tree
learning algorithms {\sc c5.0} and {\sc igtree} are mostly worse than
those of {\sc ib1-ig} on this type of data set can be further explained
by their properties.  Interpreted as statistical backed-off estimators
of the class probability given the feature-value vector, due to the
way the information-theoretic splitting criterion works, some schemata
(sets of partially matching instances) are not accessible for
extrapolation in decision tree learning. Given the high disjunctivity
of categories in language learning, abstracting away from these
schemata and not using them for extrapolation is harmful.  This type
of abstraction takes place even when no pruning is used.  Apparently,
the assumption in decision tree learning that differences in relative
importance of features can always be exploited is, for the tasks
studied, untrue. Memory-based learning, on the other hand, because it
implicitly keeps all schemes available for extrapolation, can use the
advantages of information-theoretic feature relevance weighting
without the disadvantages of losing relevant information. We plan to
expand on the encouraging results on other data sets using {\sc
tribl}, a hybrid of {\sc igtree} and {\sc ib1-ig} that leaves schemas
accesible when there is no clear feature-relevance
distinction~\cite{Daelemans+97d}.

When decision trees are pruned, implying further abstraction from the
training data, low-frequency instances with deviating classifications
constitute the first information to be removed from memory.  When the
data representing a task is highly disjunctive, and instances do not
represent noise but simply low-frequency instances that may (and do)
reoccur in test data, as is especially the case with the {\sc gs}
task, pruning is harmful to generalization.  The first reason for
decision-tree learning to be harmful (accesability of schemata) is the
most serious one, since it suggests that there is no parameter setting
that may help {\sc c5.0} and similar algorithms in surpassing or
equaling the performance of {\sc ib1-ig} in these tasks.  The second
reason (pruning), less important than the first, only applies to
data sets with low noise.  However, there exist variations of decision
tree learning that may not suffer from these problems (e.g., the lazy
decision trees of \namecite{Friedman+96}) and that remain to be
investigated in the context of our data.

Taken together, the empirical results of our research strongly suggest 
that keeping full memory of all training instances is at all times a 
good idea in language learning.


\subsection*{Acknowledgements}

This research was done in the context of the 
``Induction of Linguistic Knowledge'' research programme, supported 
partially by the Foundation for Language Speech and Logic (TSL), which 
is funded by the Netherlands Organization for Scientific Research 
(NWO).  AvdB performed part of his work at the Department of Computer 
Science of the Universiteit Maastricht.  The authors wish to thank 
Jorn Veenstra for his earlier work on the PP attachment and NP 
chunking data sets, and the other members of the Tilburg ILK group, 
Ton Weijters, Eric Postma, Jaap van den Herik, and the MLJ reviewers 
for valuable discussions and comments.

\bibliographystyle{fullname}
\bibliography{harmful-preprint}

\end{document}